\documentclass[letterpaper, 10 pt, conference]{ieeeconf}  % Comment this line out if you need a4paper

\IEEEoverridecommandlockouts                              % This command is only needed if 
\usepackage{graphics} % for pdf, bitmapped graphics files

\title{\LARGE \bf
Dynamic UGV-UAV Cooperative Path Planning \\in Uncertain Environments
}

\author{}
\author{Ninh Nguyen and Srinivas Akella% <-this % stops a space
\thanks{This work was supported in part by UNC Charlotte. The authors are with the Department of Computer Science, University of
        North Carolina at Charlotte, NC 28223, USA.
        Email: {\tt\small \{tnguy248, sakella\}@charlotte.edu}.
        }
}
\usepackage{xcolor}
\usepackage{color}

\newcommand{\saq}[1]{\textcolor{black}{#1}} % SA question
\newcommand{\nn}[1]{\textcolor{black}{#1}}
\RequirePackage[normalem]{ulem} % for \sout{}

\definecolor{boldcolor}{gray}{0.3} % range from [0,1]

\usepackage{float}
\usepackage[caption = false]{subfig}
\usepackage[final]{graphicx}
\usepackage{textcomp}

\usepackage{tabularx}

\usepackage{algorithm}
\usepackage{algpseudocode}

\usepackage{comment}
\usepackage{amsfonts}
\usepackage{amsmath}
\begin{document}

\maketitle
\pagestyle{empty}

%%%%%%%%%%%%%%%%%%%%%%%%%%%%%%%%%%%%%%%%%%%%%%%%%%%%%%%%%%%%%%%%%%%%%%%%%%%%%%%%
\begin{abstract}
This paper addresses the Dynamic UGV-UAV Cooperative Path Planning (DUCPP) problem involving one unmanned ground vehicle (UGV) assisted by one or more unmanned aerial vehicles (UAVs) operating on an uncertain road network with potentially impassable edges. DUCPP is particularly relevant for scenarios such as disaster response, emergency supply transport, and rescue operations, where a UGV must reach a specified destination in the presence of partially unknown road conditions. To enable the UGV to travel safely and efficiently to its destination, the UAV(s) dynamically inspect edges in the environment to identify and prune damaged or impassable edges from consideration.

We present multiple strategies, including a bidirectional approach, to optimize UGV-UAV cooperation for finding a safe path in an uncertain road network. Furthermore, we explore the impact of using multiple UAVs on reducing the UGV’s travel time, and evaluate the associated computation time. The proposed strategies are implemented and evaluated on 100 urban road networks. The results demonstrate that the bidirectional strategy achieves the best performance in most instances, and using multiple UAVs further reduces UGV travel time at the expense of increased computation time. This paper presents a robust framework for DUCPP to achieve efficient UGV-UAV cooperation for path planning and inspection, offering practical solutions for navigation in challenging and uncertain conditions.
\end{abstract}

%%%%%%%%%%%%%%%%%%%%%%%%%%%%%%%%%%%%%%%%%%%%%%%%%%%%%%%%%%%%%%%%%%%%%%%%%%%%%%%%
\section{Introduction}

% hurricanes, landslides  % and reliable 
% \saq{fast}
This paper presents autonomous planners for an unmanned ground vehicle (UGV) aided by unmanned aerial vehicles (UAVs) to identify safe paths through road networks obstructed or damaged  due to natural disasters (e.g., floods, earthquakes,  wildfires). The planners can also be used to suggest routes to human drivers. The goal is to perform path inspection to find safe and efficient routes for first responders, people seeking safe locations, and for trucks carrying emergency supplies (e.g., food, medicines) by adaptively updating paths in response to online sensor information.
%%reconnaissance

%\sa{However, blocked or impassable paths can cause UGVs (and even manned vehicles) to experience severe delays in operations. Tasks require efficient path planning through partially known road networks under changing conditions.}

%%\saq{search-and-rescue missions}
%Autonomous robotic systems have become indispensable in addressing challenges in dynamic and uncertain environments. Tasks such as disaster response, road safety inspection, and emergency responder transport require efficient path planning through partially known road networks under changing conditions. Unmanned ground vehicles (UGVs) can be beneficially deployed in such scenarios due to their ability to autonomously traverse carry payloads. However, when UGVs (and even manned vehicles) encounter blocked or impassable paths (e.g., due to flooding, landslides, wildfires), they can experience severe delays in operations that compromise mission success.

Since blocked or impassable paths can cause UGVs and even manned vehicles to experience severe delays, it is critical to develop techniques for efficient path finding through partially uncertain road networks whose edge statuses are gradually determined.
%To overcome these limitations, 
To achieve this, we can use the complementary abilities of unmanned aerial vehicles (UAVs). UAVs can perform aerial inspections to identify impassable paths and provide real-time updates to the UGV, enabling dynamic rerouting and safe \saq{path finding}. This cooperation reduces delays and improves operational efficiency. While progress has recently been made in utilizing a single UAV to assist a UGV, incorporating multiple UAVs provides an opportunity to accelerate inspections and improve overall performance. However, this introduces additional challenges in coordination and edge inspection assignment.
%, and communication
%\cite{BhadoriyaRDCM23, BhadoriyaMRDCM24, ChenLQCXGC24}

\subsection{Problem Context and Motivation}
In real-world applications such as disaster response, a UGV (or manned vehicle) must navigate through a road network represented as a graph to reach a target destination. UAVs can be deployed to inspect edges of this graph to identify damaged or impassable edges. This information is used by the UGV to dynamically adjust its route and avoid delays. The cooperation between the UGV and UAVs is particularly valuable in scenarios where road conditions are uncertain and impassable edges may disrupt UGV operations.

The key challenge lies in effectively coordinating UAV edge inspections to maximize their utility for UGV path finding. UAVs must prioritize edges based on their potential impact on the UGV's path. In settings with multiple UAVs, efficient task allocation is required to avoid redundant inspections and minimize computational overhead. Addressing these challenges is essential for achieving effective UGV-UAV cooperation and efficient and safe UGV path finding.

\subsection{Contributions}
% This paper addresses the Dynamic UGV-UAV Cooperative Path Planning (DUCPP) problem, which focuses on enabling efficient and safe path finding for a UGV in a graph-based environment with uncertain and potentially impassable edges. The contributions of this paper are as follows: First, we formalize the DUCPP problem by defining a framework for UGV-UAV cooperative inspection so a UGV can efficiently reach a specified destination in uncertain environments where some edges may be impassable. Second, we present and evaluate multiple strategies, including a bidirectional inspection approach, to minimize UGV travel time by optimizing UAV inspections. Third, we extend these strategies to inspection by multiple UAVs, analyzing the trade-offs between reduced travel time and computational complexity. Finally, we evaluate the proposed methods across diverse road network instances, demonstrating significant improvements in efficiency and effectiveness over baseline approaches.

This paper addresses the Dynamic UGV-UAV Cooperative Path Planning (DUCPP) problem, which focuses on using UAVs to enable efficient and safe path finding for a UGV  in a graph-based environment with uncertain and potentially impassable edges. Our main contributions are:
\begin{itemize}
    \item \textit{Problem formulation.} We formalize DUCPP to allow impassable edges whose status is revealed only when vehicles reach the damaged point, and with UAV motion that can follow graph edges for inspection or deadhead (i.e., fly directly) between vertices.
    \item \textit{Replanning strategies.} We develop several inspection and replanning strategies, including a bidirectional algorithm, and show how UAV edge inspection choices can reduce the UGV's travel time.
    \item \textit{Multiple-UAV extension.} We extend the bidirectional strategy to multiple UAVs and evaluate the reduced UGV travel time and increased computation time.
    \item \textit{Evaluation.} We perform extensive benchmarking across diverse road network instances, demonstrating significant improvements in efficiency and effectiveness over baseline approaches.
\end{itemize}

\subsection{Related Work}

Prior work has addressed replanning routes for single robots on networks with unknown obstacles~\cite{KoenigL02,PapadimitriouY91_BasicCTP}, inspecting static road networks using one or more UAVs~\cite{DilleS13, AgarwalA24}, and using UAVs to gather traffic flow information to dynamically improve resource transportation~\cite{KashyapGMSD19}.
UGV-UAV cooperation has been widely studied for 
% \sa{applications such as} 
environmental monitoring, disaster response, and agriculture~\cite{MunasinghePD24, LiuZS22}.

\subsubsection{Canadian Traveller Problem (CTP) and Variants}
When planning optimal paths for a vehicle in uncertain road environments, the Canadian Traveller Problem (CTP)~\cite{PapadimitriouY91_BasicCTP} has been used to model routing tasks where the traversability of edges is initially unknown and gradually discovered as the vehicle traverses the road network. It is PSPACE-complete to achieve a bounded competitive ratio for the CTP, and the stochastic version has been shown to be \#P-hard.
The CTP assumes that at a vertex, the vehicle can sense the traversability status of all incident edges. In contrast, our DUCPP problem assumes that the robot (UAV or UGV) only learns about the condition of the road when it reaches the \textit{damaged point}, which could be in the middle of a long road segment. This realistic modeling in DUCPP reflects the challenges faced in real-world uncertain environments where costs and road conditions are often unknown until encountered.\\
Variants of the CTP, such as $k$-CTP, have explored problems involving multiple paths, but %these approaches 
typically focus on a single UGV~\cite{BarNoyS91_StochasticCTP, BampisEX0922_kCTP, BnayaFS09_CTPRemoteSensing}. CTP with remote sensing~\cite{BnayaFS09_CTPRemoteSensing} assumes a sensor can query the state of any edge for a known cost; in DUCPP, the UAV requires path planning and does not have a fixed cost. Our work extends this idea by proposing  cooperative strategies where UAVs assist a UGV in dynamically selecting candidate paths to reach its goal in the shortest time, with the added complexity of planning for multiple agents.

\subsubsection{UGV-UAV Coordination on Road Networks}

%~\cite{ChoudhurySKP22}.
%for road networks
Much work on UGV-UAV coordination has focused on package delivery using drones and trucks~\cite{MacrinaPGL20,ChoudhurySKP22}, and assumes known road networks. Recent work~\cite{OttoGLLPR2024} has studied truck and drone delivery in road networks damaged by natural disasters. It  focuses on competitive ratio analysis of  policies where either the truck or the drone can complete the delivery. It does not provide an explicit edge inspection/replanning strategy to minimize the UGV’s travel time or benchmark the policies on realistic road-network instances.

%Most existing research in UGV-UAV coordination for road networks assumes \saq{deterministic edge conditions, or passable edges with varying costs~\cite{BhadoriyaRDCM23, BhadoriyaMRDCM24}} rather than scenarios where edges may be entirely impassable.

The most closely related research in UGV-UAV coordination for road networks assumes passable edges with varying costs~\cite{BhadoriyaRDCM23, BhadoriyaMRDCM24} rather than scenarios where edges may be entirely impassable.
In \cite{BhadoriyaRDCM23}, they proposed a framework for UGV-UAV assisted path planning in stochastic networks, where UAVs inspect edges along the UGV's $k$-shortest paths to identify delays. Their model assumes that all edges remain passable but may incur higher traversal costs depending on road conditions (e.g., congestion). They model edge cost variability, with upper and lower limits on the edge costs. However, it differs from our problem in crucial ways. 
First, once the UGV decides to traverse an edge, it cannot reverse direction. Second, their approach relies on the UGV’s expected travel time to compute the shortest path before any impeded edges are inspected. If the cost upper limit is set to infinity to model a damaged edge, the UGV fails to find a viable path since the expected travel time is infinite; using a large constant biases it to selection of paths with fewer impeded edges even though the paths may be longer. 
In contrast, in DUCPP, the UGV reroutes dynamically if its current edge is damaged — it returns to the previous vertex and replans its path.
In \cite{BhadoriyaMRDCM24}, they studied assistive routing where a support vehicle can clear impeded road segments to assist a primary vehicle, with an attendant cost.
%It assumes roads may be impeded, but are passable with higher cost.
%The most closely related work is that of 

UAV-assisted UGV routing has been explored~\cite{KashyapGMSD19} in flood scenarios, where UAVs inspect roads to identify flooded or blocked conditions. Their approach optimizes traffic flow for multiple UGVs but does not address real-time task reallocation for UAVs or dynamically prioritize inspections based on UGV needs.

% \saq{Papilloud and Keiler~\cite{PapilloudK21} and Jana et al.~\cite{JanaMNT23} focused on critical edge detection and prioritization in uncertain networks. These studies introduced graph-based techniques and metrics to identify edges that significantly impact traversal efficiency. However, these methods are primarily offline and do not incorporate real-time UGV-UAV cooperation. In contrast, our approach integrates dynamic edge prioritization and real-time replanning into UGV-UAV coordination, making it suitable for operational scenarios requiring immediate decision-making.}

% \saq{Zhou ~\cite{ZhouT20, ZhouT22} and Tokekar ~\cite{TokekarHMI16, tokekar2013sensor} proposed risk-aware path planning for UGV-UAV cooperation, where UAVs assist UGVs in navigating uncertain environments by avoiding high-risk paths. Similarly, Choudhury et al.~\cite{ChoudhurySKP22} examined UGV-UAV cooperation in deterministic applications like precision agriculture. While these studies provide valuable insights into coordination and task allocation, they are limited to environments where edges remain passable, and their focus is on deterministic or probabilistic cost models.}

A few papers~\cite{LakasBBSA18}, \cite{ZhangWHZL18},  \cite{ChenLQCXGC24} have explored vision-based strategies for UAVs to assist UGVs in navigation. 
%In these approaches, 
The UAVs capture aerial images of the environment, detect obstacles, generate and update maps in real-time, and continuously track the UGV’s position to guide it toward the destination. \saq{In contrast, our paper assumes a road network with uncertain connectivity and focuses on prioritizing critical edges for UAV inspection.}
%This approach ensures that UAVs can effectively support UGVs in dynamically adapting to updated road networks, enabling the UGV to efficiently identify and follow the optimal path to its destination.

%While promising, this strategy faces significant limitations in complex environments. For example, in areas with dense tree cover or tall buildings, the UAV's view may be obscured if it flies too high, making it challenging to detect obstacles or update maps accurately. Conversely, if the UAV operates at low altitudes, its field of view becomes constrained, preventing it from inspecting the environment comprehensively within a reasonable time. \saq{In contrast, our paper overcomes these challenges by focusing on identifying and prioritizing critical edges for UAV inspection.}

% No existing work explicitly addresses UGV-UAV cooperation in road network environments with impassable edges. The assumption in prior work that all edges are traversable, albeit with varying costs, simplifies the problem and overlooks the critical challenges posed by dynamic and uncertain edge conditions. 

%To the best of our knowledge
\nn{As described above, prior UGV-UAV work on road networks has largely emphasized edge cost uncertainty (e.g., delays) rather than connectivity uncertainty due to impassable edges discovered during execution. This paper moves toward that setting by introducing strategies that explicitly handle impassable edges and dynamically prioritize inspections to reduce UGV detours.} Additionally, it extends these strategies to multiple UAVs, providing scalable solutions to balance inspection efficiency with computational cost. %complexity.

%\nn{On delivery policies for a truck-and-drone tandem in disaster relief\cite{OttoGLLPR2024}: Similar to DUCPP, this paper studies the truck and drone cooperation; and road damage is revealed during execution, and once an edge is classified as safe or damaged, its status is assumed to remain unchanged. In contrast, their paper focuses on delivery policies where either the truck or the drone can complete the task, and they do not provide an explicit edge inspection/replanning strategy to minimize the UGV’s travel time or benchmark the policies on realistic road-network instances.}

%\nn{Coordinated Multi-Agent Pathfinding for Drones and Trucks over Road Networks \cite{ChoudhurySKP22}: Similar to DUCPP, this paper considers the UGV-UAV cooperation problem, and can also work for multiple UAVs. In contrast, it assumes the graph is known and traversable, and focuses on an offline plan that is computed in advance without dynamic update or replanning when new road information is revealed.}

\subsection{Organization}
The remainder of this paper is organized as follows. Section~\ref{sec:problem} formally states the problem and defines the key concepts. Section~\ref{sec:methods} describes the proposed UGV-UAV coordination strategies and algorithms. Section~\ref{sec:results} details the simulation experiments, evaluation metrics, and performance of the strategies. 
Section~\ref{sec:limit} addresses the practical limitations of our current approach. 
Finally, Section~\ref{sec:conclusion} summarizes the paper and discusses future research directions.

\section{Problem Statement}
\label{sec:problem}

We now state the \textit{Dynamic UGV-UAV Cooperative Path Planning (DUCPP)} problem and introduce the relevant notation. Let $G = (V, E)$ be a connected, undirected graph, where $V$ represents the set of vertices, and $E$ is the set of edges connecting these vertices. The graph models the road network the UGV must traverse to reach its destination, with potential obstacles or damage on some edges.

The UGV and UAV start at vertices $s_g$ and $s_a$ respectively, where $s_g, s_a \in V$ and $s_g$ and $s_a$ need not be identical. The UGV is tasked with reaching its selected destination vertex $d \in V$. In contrast, the UAV has no predefined destination and may terminate its journey at any vertex. Let $v_g$ and $v_a$ denote the speeds of the UGV and UAV, respectively. Each edge $e \in E$ has a length $c_e$, representing the distance of the edge. The time required for the UGV and UAV to traverse edge $e$ is given by $c_e/v_g$ and $c_e/v_a$, respectively. For simplicity, we use \textit{cost} and \textit{time} interchangeably. % \sa{in this paper}.

%\paragraph{UGV and UAV motion models}
The UGV is constrained to move along the edges of the graph $G$. The UAV can move along an edge of $G$ to inspect it, or it can \textit{deadhead} from its current position directly to any vertex $v \in V$ in straight-line flight, with travel time given by the Euclidean distance divided by its speed. For example, suppose the UGV's current path is blocked (and detected by either UGV or UAV). In this case, a new path is computed for the UGV, and the UAV may be reassigned to fly directly to a vertex of an uninspected edge of that path to inspect. This formulation captures the UAV's key advantage: its ability to move flexibly through free space to support the UGV's navigation task.

An edge is initially labeled \textit{uninspected}. After traversal by a UAV or UGV, the edge is labeled as \textit{safe} if it is passable or \textit{damaged} if it is impassable. We define a \textit{damaged edge} or an \textit{impassable edge} as an edge with damage or an obstacle so that the UGV cannot traverse it. Let $D \subseteq E$ be the set of damaged edges in the graph that the UGV cannot traverse. The set of damaged edges is initially unknown and can only be determined gradually when either a UAV or the UGV reaches a damaged point or obstacle on an edge. If the UGV encounters a damaged edge during traversal, it must return to the last visited vertex and compute a new path. In such cases, the damaged edge is removed from the graph.
%\sout{, and its cost is set to $c_e = \infty$. For the UAV, the traversal cost remains unchanged regardless of the edge's condition.} \saq{(So should we have separate edge costs for the UGV and UAV?)}

\textit{Objective:}
The objective is to minimize the total travel time $C/v_g$ for the UGV to reach its destination vertex $d$, where $C$ represents the total travel distance of the UGV. If no path exists from the starting vertex $s_g$ to the destination vertex $d$ due to damaged edges, then $C = \infty$. This paper proposes strategies to enable the UGV to reach its destination in the shortest possible time by leveraging UAV inspections to identify and eliminate damaged edges effectively.

% \paragraph{Assumptions}
We make the following assumptions to isolate the \textit{planning} component of DUCPP: (i) all vehicles start simultaneously, (ii) edge status updates are communicated reliably and without delay, (iii) replanning time is negligible relative to edge traversal/inspection time, (iv) UAVs can deadhead in straight lines from their position to any vertices, (v) UAVs have no energy constraints, and (vi) once inspected, an edge's status remains constant.
% \sa{Section~\ref{sec:limit} discusses how energy limits, communication latency, and sensing uncertainty can be incorporated.}

% We will make the following assumptions: (1) all vehicles start their travel simultaneously, (2) the UGV and UAV(s) can communicate information without delay, (3) the robots can calculate a new route and change their direction instantaneously, (4) the UAV can travel directly in a straight line from its current position to any vertex in the graph without following edges, (5) once an edge is inspected, its condition (damaged or safe) remains constant throughout the problem.

\section{Algorithms for DUCPP}
%\saq{Inspection Strategies?}
\label{sec:methods}
Efficient coordinated inspection between the UGV and UAV(s) requires a framework that facilitates communication between the robots, rapid path recomputation in response to sensor information, and effective edge inspection assignment to minimize the total travel time of the UGV. To achieve this, the high-level strategy  is divided into the following steps:
\begin{enumerate}
    \item \textbf{Path Planning:} Determine the best possible path for the UGV to reach its destination based on the current state of the graph.
    \item \textbf{Edge Assignment:} Identify critical edges from the UGV's current path and assign them to the UAV(s) for inspection. We assume each UAV is assigned one edge at a time.
    \item \saq{\textbf{Path Traversal and Edge Inspection:}} %% Task Execution
    The UGV and UAV(s) follow their assigned paths and update the \textit{edge status} (safe or damaged) upon completing an edge traversal or encountering an obstacle.
\end{enumerate}
These steps are repeatedly executed whenever the graph is updated, continuing until either the UGV reaches its destination or finds that no feasible path exists.

\begin{algorithm}
    \footnotesize
    \caption{UGV-UAV Coordinated Path Planner}
    \label{alg:main}
    \begin{algorithmic}[1]
    \Procedure{Main}{$strategy$}
        % \State {init strategy}
        \State {$totalTime = 0.0$}
        \While {True}
            \State {$strategy.find\_path()$}
            \State {$stepTime, keepGoing$=$strategy.find\_event\_and\_update()$}
            \State{$totalTime$ += $stepTime$}
            
            \If {not $keepGoing$}
                \State {break}
            \EndIf
        \EndWhile
    \EndProcedure
    \end{algorithmic}
    
\end{algorithm}
As outlined in Algorithm~\ref{alg:main}, communication and coordination between the robots (UGV and UAVs) are used to iteratively achieve the solution. The \textbf{Path Planning} and \textbf{Edge Assignment} tasks are executed by each strategy at line 4, determining and assigning paths to the robots. Subsequently, line 5 handles the \saq{\textbf{Path Traversal and Edge Inspection}}, \saq{which calculates the robot positions at the next event where replanning becomes necessary.}
An \textit{event} occurs when a robot (UGV or UAV) must stop due to encountering an obstacle or damage on an edge, a UAV completes an edge inspection, or the UGV reaches its destination.

This process enables efficient dynamic replanning and ensures that both the UGV and UAV(s) adapt to the evolving graph conditions. In this paper, we present and evaluate multiple strategies within this framework, evaluating their performance based on the UGV's total travel time and computational time. The modular framework allows ease of implementation and testing of new strategies.

While the UGV and UAV exhibit different behaviors, both share primary functions:
\begin{itemize}
    \item \texttt{find\_next\_event()}: Executes the assigned path and returns the event when the vehicle stops or completes its task. If the vehicle encounters an obstacle or damage on an edge, it stops, and the affected edge is marked as a damaged edge and removed from the graph. A UAV is considered to have completed its task upon fully inspecting an edge, while the UGV completes its task upon reaching its destination.
    \item \texttt{update\_position(event)}: Recalculate the new position of each vehicle at an event; all vehicles need to stop and recalculate their new paths.
\end{itemize}

\subsection{Abstract Strategy}
\saq{The main idea for solving the DUCPP problem is to find the best path for the UGV to reach the destination and then analyze that path to assign an edge to inspect to the UAV to support the UGV.} The most important parts of the strategy are in two functions:
\begin{itemize}
    \item \texttt{find\_path()} finds the path for each robot. This differs for each strategy.
    \item \texttt{find\_event\_and\_update()} returns the elapsed travel time to the next event  on the assigned paths. Since \texttt{find\_event\_and\_update()} is the same for all strategies, it is implemented in the $AbstractStrategy$ class.
\end{itemize}

\begin{algorithm}
    \footnotesize
    \caption{Abstract Strategy Class}
    \label{alg:abstract_strategy_class}
    \begin{algorithmic}[1]
    \Procedure{find\_event\_and\_update()}{}
        \If {not $ugv.path$}
            \State{}
            \Return{$(0, False)$}
        \EndIf

        % \State{}
        \State {\# Find the next event when a robot has to stop}
        % (i.e., damaged}
        % \State { edge found, UAV completes edge inspection, UGV reaches destination)}
        \State {$uavTime = uav.find\_next\_event()$}
        \State {$ugvTime = ugv.find\_next\_event()$}
        \State {$nextEventTime = min(uavTime, ugvTime)$}
        % \State{}
        \State {$uav.update\_position(nextEventTime)$}
        \State {$ugv.update\_position(nextEventTime)$}

        \State{}
        \If {the UGV reaches the destination}
            \State{}
            \Return {$(nextEventTime, False)$}
        \EndIf

        \State{}
        \Return {$(nextEventTime, True)$}
    \EndProcedure
    \end{algorithmic}
\end{algorithm}
% \vspace{-0.2cm}

The \texttt{find\_event\_and\_update} function (Algorithm \ref{alg:abstract_strategy_class}) returns both the next event time and the $keepGoing$ boolean value. As shown in lines $2$-$4$, if no valid path exists from the UGV’s current position to the destination, the $keepGoing$ value is set to $False$. Lines $6$-$10$ compute the next event where all robots must stop and recalculate their paths. Since our primary focus in this paper is calculating UGV travel time and computation time, we utilize known information about obstacle positions for this process. However, in a real-world application, this function would need to be modified to dynamically receive event updates from the UGV and UAV during motion and inspection. Despite this difference, the results for travel time and computation time remain consistent between the two versions.

\subsection{\saq{Implemented} Strategies}
\subsubsection{Perfect knowledge} This strategy establishes a conservative \textit{lower bound} for the problem by assuming the UGV is given complete knowledge of the status of each edge (safe or damaged) in the road network. All damaged edges are removed from the graph, and Dijkstra's algorithm is used to find the shortest path to the destination $d$. If no safe path exists, the UGV does not move and so its travel time is set to 0.
\subsubsection{UGV-only} In this simple strategy, which serves as a baseline, we use only a single UGV to find a safe path to the destination. The UGV is always assigned the shortest path (computed using Dijkstra, A*, or D* Lite) from the current position to the destination. Whenever the UGV finds a damaged point on the path, the UGV must return to the last visited vertex and reroute on a new shortest path.

\subsubsection{UGV-UAV with Kemeny constant} Here we pre-compute the criticality of each edge in the graph. For this, we use a measure of centrality of an edge $e$ in an undirected graph introduced by \cite{AltafiniBCMP22}. It is a modification of the Kemeny constant of the graph; for each edge, its value is the Kemeny constant after removing the edge. The UGV always follows the shortest path, and the UAV inspects the edge with the highest critical value on the UGV's path (Algorithm~\ref{alg:kemeny_strategy}). Although the modified Kemeny constants should ideally be recomputed every time the graph is updated, for computational reasons, we compute the Kemeny constants only once at the beginning.
\begin{algorithm}
    \footnotesize
    \caption{UGV-UAV with Kemeny constant}
    \label{alg:kemeny_strategy}
    \begin{algorithmic}[1]
    \Procedure{find\_path()}{}
        \State {\# Find the shortest path and assign it to the UGV}
        \State {$ugv.path = find\_shortest\_path(ugv.position, d)$}
        \If {not $ugv.path$}
            \State {}
            \Return{$False$}
        \EndIf

        % \State{}
        \State{\# Assign the edge with the highest critical value to the UAV}
        \For {$edge$ \textbf{in} $ugv.path$}
            \If {$edge$ is not inspected \textbf{and} has higher critical value}
                \State {$uav.path = edge$  \# Edge to be inspected}
            \EndIf 
        \EndFor
    \EndProcedure
    \end{algorithmic}
\end{algorithm}

\subsubsection{UGV-UAV with k-shortest paths} This strategy (Algorithm~\ref{alg:k_paths_strategy}) uses a different approach to identify the critical edge to be inspected by the UAV. Each time the UGV reroutes, the UAV  uses Yen's algorithm~\cite{Yen71} to find the $k$ shortest paths from the current UGV position to the destination. It then inspects the edge that appears most frequently in the $k$ shortest paths, provided the edge is also in the UGV's current path. 
\begin{algorithm}
    \footnotesize
    \caption{UGV-UAV with $k$-shortest paths}
    \label{alg:k_paths_strategy}
    \begin{algorithmic}[1]
    \Procedure{find\_path()}{}
        \State {\# Find the shortest path and assign it to the UGV}
        \State {$ugv.path = find\_shortest\_path(ugv.position, d)$}
        \If {not $ugv.path$}
            \State {}
            \Return {$False$}
        \EndIf

        % \State{}
        \State{\# Assign the edge that is most repeated in the $k$-shortest paths}
        \State{$kPaths = find\_k\_paths(ugv.position, d)$}
        \State{Assign uninspected edge that is most repeated in $kPaths$ to UAV}
        
    \EndProcedure
    \end{algorithmic}
\end{algorithm}

\subsubsection{UGV-UAV with MPSP} The Most Probable Shortest Path (MPSP) problem in a graph aims to find the path that not only minimizes the cost but also maximizes the probability of successful traversal. This approach is particularly relevant in uncertain environments where edges have \textit{existence probabilities}, i.e., probabilities representing their reliability or likelihood of being passable. In Algorithm~\ref{alg:MPSP_strategy}, we use the algorithm in \cite{SahaBVKB21} to find the MPSP and assign that path to the UGV. Then, the UAV inspects the edge with the lowest existence probability (i.e., least likely to be passable).
\begin{algorithm}
    \footnotesize
    \caption{UGV-UAV with Most Probable Shortest Path}
    \label{alg:MPSP_strategy}
    \begin{algorithmic}[1]
    \Procedure{find\_path()}{}
        \State {\# Find the MPSP and assign it to the UGV}
        \State {$ugv.path = find\_MPSP\_path(ugv.position, d)$}
        \If {not $ugv.path$}
            \State {}
            \Return {$False$}
        \EndIf

        % \State{}
        \State{\# Assign the edge with the lowest existence probability to the UAV}
        \For {$edge$ \textbf{in} $ugv.path$}
            \If {$edge$ not inspected \textbf{and} has lower existence probability}
                \State {$uav.path = edge$ \# Edge to be inspected}
            \EndIf
        \EndFor
    \EndProcedure
    \end{algorithmic}
\end{algorithm}

\subsubsection{UGV-UAV with bidirectional approach}
In bidirectional search, one search tree grows from the start node while another grows from the destination node, and a path is found when the two trees meet~\cite{LaValle06}. For many problems, bidirectional search significantly reduces the required exploration. Variants such as bidirectional Dijkstra and A* provide optimal solutions when positive edge costs or an admissible heuristic function are available. For dynamic networks, \cite{LiMWM23} proposed advanced bidirectional algorithms like Bidirectional Lifelong Planning A* (BLPA*) and fractional bidirectional D* Lite (fBD* Lite(dp)), which efficiently recompute shortest paths in response to environmental changes. However, these algorithms are designed for single-agent scenarios and do not address \saq{task assignment or} inspection planning by supporting agents, such as UAVs. This limits their suitability for scenarios requiring multi-agent cooperation to handle dynamic, uncertain environments.
\begin{algorithm}
    \footnotesize
    \caption{UGV-UAV with Bidirectional Strategy}
    \label{alg:bidirectional_strategy}
    \begin{algorithmic}[1]
    \Procedure{find\_path()}{}
        
        % \State {$ugv.path = None$}
        % \State {$lastNode, nextNode, ratio = ugv.relativePosition$}

        % \State
        % \State {\textit{\# UGV needs to come back to the last safe node}}
        % \If {$edge(lastNode, nextNode)$ was removed}
        %     \State {$lastNode, nextNode, ratio = nextNode, lastNode, 1 - ratio$}
        % \EndIf

        % \State
        % \State {\textit{\# Find the shortest path from the UGV to the destination}}
        % \State {$path, cost = find\_shortest\_path(nextNode, d)$}
        % \If {$lastNode != nextNode$ and $G.hasEdge(lastNode, nextNode)$}
        %     \State {$path2, cost2 = find\_shortest\_path(lastNode, d)$}
        %     \If {$cost2 + distance(lastNode, nextNode) * ratio < $
        %         $\;\;\;\;cost + distance(nextNode, lastNode) * (1-ratio)$}
        %         \State {$path, cost = path2, cost2$}
        %     \EndIf
        % \EndIf
            
        % \State
        % \State {\textit{\# Return False if cannot find any paths}}
        % \If {$path$ not found}
        %     \State {$return False$}
        % \EndIf
        % \State {$ugv.path = path$}

        \State {\# Find the shortest path and assign it to the UGV}
        \State {$ugv.path = find\_shortest\_path(ugv.position, d)$}
        \If {not $ugv.path$}
            \State {}
            \Return{$False$}
        \EndIf

        % \State
        \State \textit{\# For UAV: Inspect the next uninspected edge on reversed UGV path}
        \State {$uav.path = None$}
        \For {$i \gets length(ugv.path)-1$ to $1$}
            \If {$edge(ugv.path[i], ugv.path[i-1])$ not inspected}
                \State {$uav.path = [ugv.path[i], ugv.path[i-1]]$}
                \State {break}
            \EndIf
        \EndFor
    \EndProcedure
    \end{algorithmic}
\end{algorithm}
Our work extends the bidirectional approach to address these challenges, adapting it for inspection by UAVs in UGV-UAV cooperative scenarios.  We compute the shortest path from the UGV's current position to the destination and simultaneously assign the UAV to inspect the reversed path from the destination back toward the UGV (Algorithm~\ref{alg:bidirectional_strategy}). This strategy ensures the UAV can quickly inspect edges along the UGV's path \saq{without unnecessary UAV transit travel or redundant inspections}. If either robot encounters an impassable edge, the affected edge is removed from the graph, and both robots recalculate their respective paths. This process repeats iteratively until either the UGV successfully reaches its destination or no viable path remains.

\begin{figure}[!ht]
\centering
\subfloat[Event 1]{\includegraphics[width=0.8\linewidth]{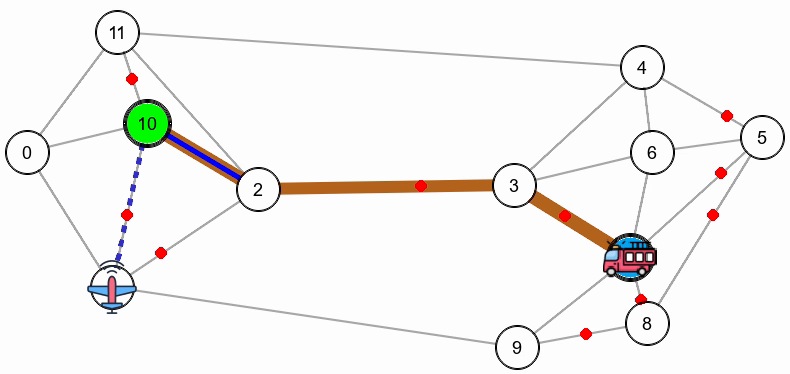}} \label{fig:1_a}\\
% \vspace{-0.3cm}
\subfloat[Event 2]{\includegraphics[width=0.8\linewidth]{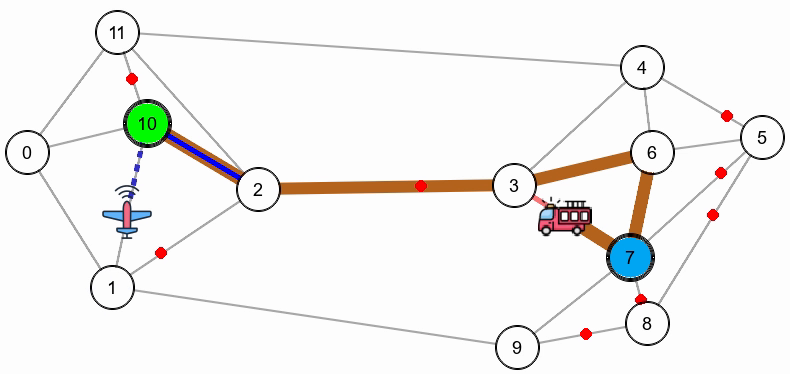}}
\caption{(a) An example road network showing the start positions of the UGV and UAV and the initial shortest path (in brown) from start vertex 7 to the destination vertex 10. The UGV is represented by the firetruck, and the UAV by the plane. (b) The paths and positions of robots at event 2, when the UGV encounters an obstacle.  The red dots represent (initially unknown) obstacles or obstructions on edge segments.
}
\label{fig:bidirectional_1}
\end{figure}
% \vspace{-0.3cm}

%In contrast to the strategies that assign edges for inspection to the UAV based on edge priority metrics (e.g., criticality or appearance frequency in shortest paths), our bidirectional approach emphasizes efficiency in UAV travel. Other strategies may require the UAV to travel long distances to reach high-priority edges. 
Our bidirectional approach emphasizes efficiency in UAV travel. This is in contrast to strategies that assign edges for inspection to the UAV based on edge priority metrics (e.g., criticality or appearance frequency in shortest paths); these may require the UAV to travel long distances to reach high-priority edges.
In the bidirectional strategy, the UAV inspects uninspected edges in reverse order along the UGV’s shortest path, typically starting with the edges closest to the destination. This reduces UAV travel time significantly and, in many cases, allows the UAV to inspect consecutive edges of the UGV path without interruption, as these edges are usually connected. 
%% Might not be connected if the path includes safe edges.
The bidirectional approach is also more computationally efficient.

% \sout{Line $3$ returns the current position of the UGV with the current edge $(lastNode, nextNode)$ and the ratio between the traversed distance and the total length of the edge. It means $ratio = 0$ when the UGV is at $lastNode$, and $ratio = 1$ when the UGV is at $nextNode$. From that information, we can find the shortest path from the current position of the UGV to the destination in code lines $10-17$. After that, the strategy will identify the uninspected edge in the backward order of the UGV's path and assign it to the UAV for inspection.}

\begin{figure}[!ht]
\subfloat[Event 3]{\includegraphics[width=0.5\linewidth]{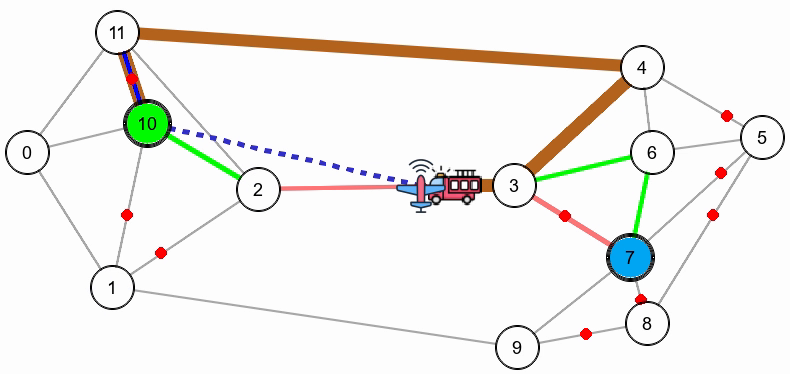}}
\subfloat[Event 4]{\includegraphics[width=0.5\linewidth]{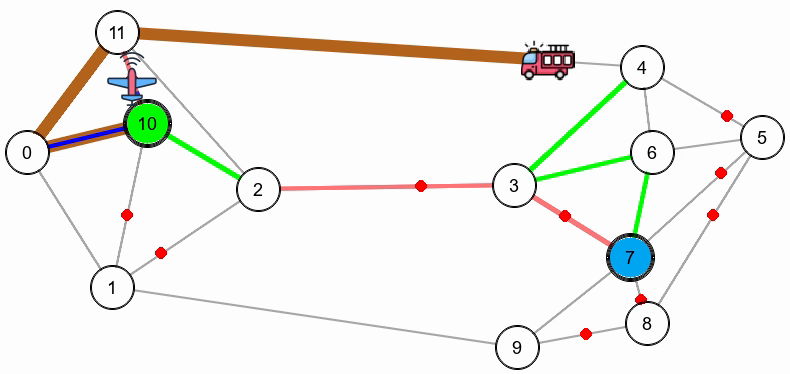}}\\
\subfloat[Event 5]{\includegraphics[width=0.5\linewidth]{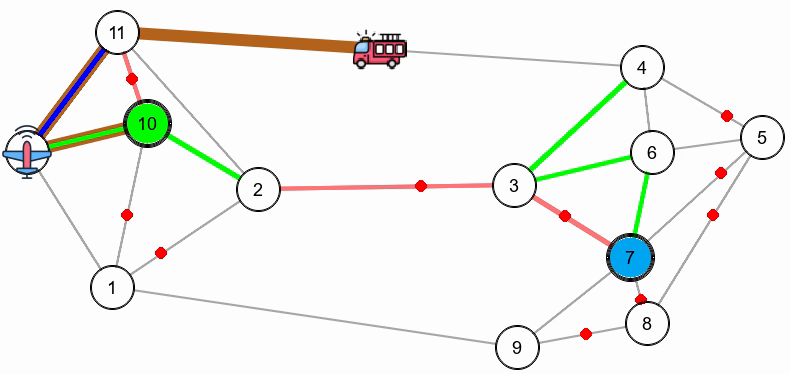}}
\subfloat[Event 6]{\includegraphics[width=0.5\linewidth]{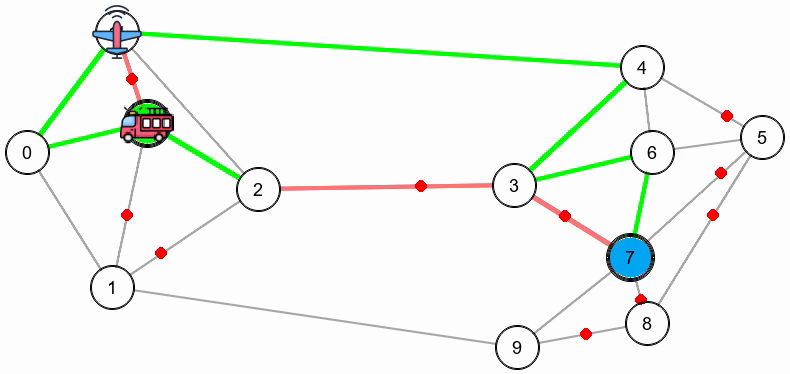}}
\caption{The paths and positions of robots at selected events.
The UGV arrives at the destination (event 6) after several edges have been inspected by the UGV and UAV. Safe edges are shown in green, damaged edges in red. The final safe path traversed by the UGV is $\langle 7, 6, 3, 4, 11, 0, 10 \rangle$.}
\label{fig:bidirectional_2}
\end{figure}

We illustrate the bidirectional strategy with a simple example (Figures~\ref{fig:bidirectional_1} and~\ref{fig:bidirectional_2}).
In this example, the UAV starts at node $1$, the UGV starts at node $7$, and the UGV's destination is node $10$. At the beginning event (Fig. 1(a)), the UGV's shortest path from its start to the destination is $\langle 7, 3, 2, 10 \rangle$  \saq{(highlighted in brown)}. Since the statuses of edges in the map are unknown, the UAV will try to inspect that path in reverse order. Hence, the first edge that the UAV needs to inspect is $(10, 2)$. The robots begin moving after the paths are assigned to them. Because there is a damaged point in the middle of the edge $(7, 3)$, the UGV will reach and find it before the UAV can complete inspecting the edge $(10, 2)$. Event 2 shows the positions of the robots and their planned paths after the UGV finds a damaged point. When any robot finds a damaged point, the graph is updated, and all robots recalculate their new paths. The UGV needs to return to the last safe node, and the UGV's updated path will be $\langle 7, 6, 3, 2, 10 \rangle$. The UAV's path is still $(10, 2)$.
% $7 \rightarrow 3 \to 2 \rightarrow 10$
% $10 \rightarrow 2$
% $7 \rightarrow 6 \rightarrow 3 \rightarrow 2 \rightarrow 10$
% $10 \rightarrow 2$

% \vspace{-0.4cm}

\begin{itemize}
    \item \textbf{Event 3:} The UAV completed the inspection of the edge $(10, 2)$ and marked it as a safe edge, then continued to inspect the edge $(2, 3)$. The UGV found a damaged point first on edge $(2, 3)$ and stopped; the UAV received the information from the UGV and stopped on edge $(2, 3)$ where it was inspecting it. The next UGV path is $\langle 3, 4, 11, 10 \rangle$, and the UAV is assigned edge $(10, 11)$.  %$3 \rightarrow 4 \rightarrow 11 \rightarrow 10$
    \item \textbf{Event 4:} The UAV found a damaged point on the edge $(10, 11)$ first, while the UGV was traversing the edge $(4, 11)$. The next UGV path is $\langle 11, 0, 10\rangle$, and so the UAV's assigned edge is $(10, 0)$.  %$11 \rightarrow 0 \rightarrow 10$
    \item \textbf{Event 5:} The UAV completed inspecting the edge $(10, 0)$ when the UGV was still on the edge (4, 11). The UGV's path is $\langle 11, 0, 10 \rangle$, and the UAV's edge is $(0, 11)$.  %$11 \rightarrow 0 \rightarrow 10$
    \item \textbf{Event 6:} The UAV completed inspecting the edge $(0, 11)$. Since all edges in the UGV's current path are safe, the UAV stopped at node $11$, and the UGV continues until it reaches its destination.
\end{itemize}

The example underscores the ability of the bidirectional strategy to dynamically adapt to graph updates while efficiently coordinating the UGV and UAV.

\subsubsection{UGV and multiple UAVs with bidirectional approach}
\label{intro-multi-uavs-bidirectional}
We extended the bidirectional approach to multiple UAVs by combining it with the $k$-shortest paths algorithm. Specifically, given $k$ UAVs, we apply Yen’s algorithm~\cite{Yen71} to compute the $k$ shortest paths from the UGV's current position to the destination $d$. Each UAV is then assigned to inspect the first uninspected edge along one of these reversed paths that has not already been allocated to another UAV. This assignment is performed iteratively until either all UAVs have been allocated or no further uninspected edges remain. This procedure guarantees that no edge is assigned more than one UAV, thereby eliminating redundant inspections. (This also avoids collisions along edges. If multiple UAVs are assigned edges with a common vertex, we assume they avoid collisions by flying at different altitudes.)

% With a single UAV, the bidirectional strategy needs to compute only one shortest path for the UGV in each iteration, with a complexity of $O((V+E)\log V)$. In contrast, extending to  $k$ UAVs with Yen’s algorithm substantially increases the computational cost of each iteration to $O(kV((V+E)\log V))$, reflecting the overhead of generating and managing multiple candidate paths.

\subsection{Time Complexity}
Table \ref{table:complexity} summarizes the time complexity of \texttt{find\_path()} for each of the strategies; it does not include the preprocessing time for the Kemeny constants. The complexity depends primarily on the number of vertices ($V$) and edges ($E$) in the graph. The UGV-only and bidirectional strategies use Dijkstra's algorithm, ensuring efficient path computation. In contrast, strategies involving $k$-shortest paths and multiple UAVs introduce additional complexity due to the need for computing multiple shortest paths~\cite{Yen71}.
The MPSP strategy performs Dijkstra’s algorithm $m$ times to sample candidate paths; we use $m=20$. Additionally, its complexity is influenced by $N$, the number of Monte Carlo (MC) runs in the Luby-Karp algorithm, contributing to the overall computational cost.

\begin{table}[htbp]
\caption{Time complexity of different strategies.}
% ($V$ is the number of vertices. $E$ is the number of edges, and $k$ represents the number of shortest paths computed)
\centering
\begin{tabular}{||c c||}
\hline
Strategy & Complexity  \\ [0.5ex]
\hline
UGV-only & $O((V + E) \log V)$\\ 
Kemeny & $O((V + E) \log V)$ \\ 
$k$-Shortest Paths  & $O(kV((V + E) \log V))$\\ 
MPSP & $O(m(NE + V\log V + \log m))$ \\ 
Bidirectional & $O((V + E) \log V)$\\ \hline
\multicolumn{2}{||c||}{Multiple UAVs with Bidirectional Strategy} \\
$k$ UAVs & $O(kV((V + E) \log V))$\\ 
%3-UAVs ($k$ = 3) & $O(kV((V + E) \log V))$\\ 
%5-UAVs ($k$ = 5) & $O(kV((V + E) \log V))$\\ 
%7-UAVs ($k$ = 7) & $O(kV((V + E) \log V))$\\
 [1ex]
 \hline
\end{tabular}
\label{table:complexity}
\end{table}
% \vspace{-0.6cm}

\section{Results}
\label{sec:results}
\subsection{Datasets}
To evaluate the performance of the strategies, we utilized the Line Coverage dataset~\cite{AgarwalA24}, derived from OpenStreetMap, containing road networks from the 50 most populous cities globally. Each city has two road networks: a small map spanning an area approximately 1 km $\times$ 1 km and a large map covering 3 km $\times$ 3 km.

For each road network, 50 unique instances were generated by randomly selecting damaged edges and assigning obstacle positions on these edges using a uniform distribution. Each edge was assigned an existence probability, randomly sampled from the interval $[0.6, 1.0]$. The minimum probability was set to 0.6 to ensure that the graphs remain sufficiently connected; lower probabilities resulted in too many damaged edges, making it difficult to find a viable path from the start to the destination. To maintain consistency across strategies, each instance was generated with a unique random seed, ensuring the same graph configurations are used for testing all strategies.
The length of each edge is known, and is used to calculate the time a UGV needs to traverse it or a UAV needs to inspect it. However, when a UAV needs to reach an edge for inspection, we add the shortest straight-line UAV travel cost to the edge to perform the inspection.

The start and destination vertices of the UGV and UAV(s) were also randomly chosen for each instance.

\subsection{Evaluation Metrics}
To assess the effectiveness of the proposed strategies, we employ two evaluation metrics:

\begin{itemize}
    \item \textbf{UGV Travel Time:} This is the total time taken by the UGV to reach its destination. In cases where no viable or safe path exists from the UGV's starting position to the destination, the travel time is defined as the time from the initial start time until the program determines that no viable path is available.
    \item \textbf{Computation Time:} This is the total compute time needed by the algorithm to recalculate paths for the UGV and UAV(s) whenever new information (e.g., a damaged edge) is discovered or an inspection task is completed. It reflects the computational efficiency of the strategy and its ability to handle real-time updates.
\end{itemize}

These two metrics provide complementary insights into the performance of the strategies. While computation time highlights the responsiveness of the algorithms, travel time evaluates their overall effectiveness in guiding the UGV to its destination under uncertain conditions.

\subsection{Results}
The travel time and computation time performance of the presented strategies were analyzed across both small and large maps.
% \sa{The strategies compared include: \textit{Perfect knowledge}, \textit{UGV-only path finding}, \textit{Kemeny critical edge prioritization}, \textit{k-shortest paths inspection}, \textit{MPSP}, and \textit{bidirectional strategies}.}
A multi-UAV bidirectional strategy with 3, 5, and 7 UAVs was also explored to assess scalability.

%% 20 m/s is 44.7387 miles per hour and 72 kmph.
To ensure consistency across all scenarios, we set the UGV speed to a constant value of 20 m/s. To evaluate the impact of UAV speed on the strategies, we tested three different speeds for the UAV(s): 20 m/s, 30 m/s, and 40 m/s. These variations capture a wide range of realistic operational conditions and allow us to analyze how the relative speeds of the UGV and UAV(s) affect the overall performance of the strategies.

\begin{table}[ht!]
\scriptsize
\caption{Average UGV travel times (in seconds) for five small maps. The strategies are compared for no-UAV and single-UAV cases. 
Multiple-UAV cases use the bidirectional strategy.
}
\centering
% \hspace*{-2.5cm}
\resizebox{0.48\textwidth}{!}{
    \begin{tabular}{||c c c c c c||} 
    \hline
     & Moscow & Sao Paulo & Lagos & Tokyo & Mexico City  \\ [0.5ex]
    \hline
    \hline
    \multicolumn{6}{||c||}{Lower bound with perfect knowledge} \\
    \hline
    Perfect knowledge & 9.334 & 21.401 & 17.543 & 8.919 & 25.187\\
    \hline
    \hline
    \multicolumn{6}{||c||}{No-UAV and Single-UAV strategies} \\
    \hline
    UGV-only & 40.968 & 68.855 & 74.095 & 31.421 & 92.167\\ 
    Kemeny & 28.060 & 42.226 & 58.834 & 22.380 & 63.371\\ 
    $k$-shortest paths & 36.424 & 56.533 & 65.071 & 28.970 & 79.559\\ 
    MPSP & 28.609 & 50.164 & \textbf{49.386} & 22.286 & 73.552\\ 
    Bidirectional & \textbf{25.620} & \textbf{37.292} & 54.360 & \textbf{21.085} & \textbf{59.758}\\ 
    \hline
    \hline
    \multicolumn{6}{||c||}{Multiple UAVs with bidirectional strategy} \\
    \hline
    3 UAVs & 23.118 & 35.460 & 51.198 & 19.094 & 59.951\\ 
    5 UAVs & 22.664 & 35.824 & 50.111 & 18.230 & 57.659\\ 
    7 UAVs & \textbf{22.506} & \textbf{35.160} & \textbf{49.111} & \textbf{17.221} & \textbf{54.526}\\ 
     [1ex]
    \hline
    \end{tabular}
}
\label{table:avg-travel-small-20-40}
\end{table}
% \vspace{-0.3cm}

Due to space constraints, we only show the results for the most populous city from five continents (Africa, Asia, Europe, North America, and South America) for a UAV speed of 40 m/s. Tables~\ref{table:avg-travel-small-20-40} and  \ref{table:avg-travel-large-20-40}, and Tables~\ref{table:avg-computation-small-20-40} and \ref{table:avg-computation-large-20-40} provide the average UGV travel times and computation times, respectively, for small and large maps averaged across 50 random instances for each map. The Perfect knowledge and UGV-only strategies are included only as baselines.

\begin{table}[ht!]
\scriptsize
\caption{Average UGV travel times (in seconds) for five large maps.
The strategies are compared for no-UAV and single-UAV cases.
%in the upper part of the table. 
Multiple-UAV cases 
%in the lower part of the table 
use the bidirectional strategy.
}
\centering
\resizebox{0.48\textwidth}{!}{
    \begin{tabular}{||c c c c c c||} 
    \hline
     & Moscow & Sao Paulo & Lagos & Tokyo & Mexico City  \\ [0.5ex]
    \hline
    \hline
    \multicolumn{6}{||c||}{Lower bound with perfect knowledge} \\
    \hline
    Perfect knowledge & 31.024 & 44.070 & 48.328 & 85.778 & 98.602\\
    \hline
    \hline
    \multicolumn{6}{||c||}{No-UAV and Single-UAV strategies} \\
    \hline
    UGV-only & 135.051 & 212.231 & 317.366 & 273.283 & 364.676\\ 
    Kemeny & 85.720 & 147.122 & 216.049 & 207.027 & 273.582\\ 
    $k$-shortest paths & 116.162 & 174.797 & 262.704 & 249.550 & 320.561\\ 
    MPSP & 88.317 & \textbf{133.532} & 201.392 & 275.395 & 285.462\\ 
    Bidirectional & \textbf{78.533} & 152.844 & \textbf{170.639} & \textbf{171.723} & \textbf{196.409}\\ 
    \hline
    \hline
    \multicolumn{6}{||c||}{Multiple UAVs with bidirectional strategy} \\
    \hline
    3 UAVs & 75.907 & 117.698 & 147.994 & 164.417 & 193.672\\ 
    5 UAVs & 76.135 & 124.260 & 149.846 & 165.351 & 188.807\\ 
    7 UAVs & \textbf{71.947} & \textbf{116.775} & \textbf{138.264} & \textbf{159.461} & \textbf{183.182}\\ 
     [1ex]
    \hline
    \end{tabular}
}

\label{table:avg-travel-large-20-40}
\end{table}
% \vspace{-0.5cm}

\begin{table}[ht!]
\scriptsize
\caption{Average computation times (in seconds) for five small maps.
% The Perfect Knowledge and UGV-only strategies are included only as baselines.
}
\centering
\resizebox{0.48\textwidth}{!}{
    \begin{tabular}{||c c c c c c||} 
    \hline
     & Moscow & Sao Paulo & Lagos & Tokyo & Mexico City  \\ [0.5ex]
    \hline
    \hline
    \multicolumn{6}{||c||}{Lower bound with perfect knowledge} \\
    \hline
    Perfect knowledge & 0.009 & 0.025 & 0.037 & 0.015 & 0.042 \\
    \hline
    \hline
    \multicolumn{6}{||c||}{No-UAV and Single UAV strategies} \\
    \hline
    UGV-only & 0.010 & 0.031 & 0.059 & 0.020 & 0.108 \\ 
    Kemeny & 0.012 & 0.038 & \textbf{0.085} & 0.025 & \textbf{0.168} \\ 
    $k$-shortest paths & 0.044 & 0.229 & 1.086 & 0.168 & 3.947 \\ 
    MPSP & 0.029 & 0.182 & 0.476 & 0.093 & 1.642 \\ 
    Bidirectional & \textbf{0.011} & \textbf{0.035} & 0.111 & \textbf{0.024} & 0.233 \\ 
    \hline
    \hline
    \multicolumn{6}{||c||}{Multiple UAVs with bidirectional strategy} \\
    \hline
    3 UAVs & \textbf{0.038} & \textbf{0.166} & \textbf{1.424} & \textbf{0.126} & \textbf{4.766} \\ 
    5 UAVs & 0.075 & 0.336 & 3.085 & 0.254 & 10.700 \\ 
    7 UAVs & 0.123 & 0.509 & 4.877 & 0.372 & 15.654 \\ 
     [1ex]
    \hline
    \end{tabular}
}
\label{table:avg-computation-small-20-40}
\end{table}
% \vspace{-0.3cm}

\begin{table}[ht!]
\scriptsize
\caption{Average computation times (in seconds) for five large maps. 
% The Perfect Knowledge and UGV-only strategies are included only as baselines.
}
\centering
\resizebox{0.48\textwidth}{!}{
    \begin{tabular}{||c c c c c c||} 
    \hline
     & Moscow & Sao Paulo & Lagos & Tokyo & Mexico City  \\ [0.5ex]
    \hline
    \hline
    \multicolumn{6}{||c||}{Lower bound with perfect knowledge} \\
    \hline
    Perfect knowledge & 0.095 & 0.216 & 0.188 & 0.341 & 0.391\\
    \hline
    \hline
    \multicolumn{6}{||c||}{No-UAV and Single UAV strategies} \\
    \hline
    UGV-only & 0.098 & 0.238 & 0.221 & 0.452 & 0.513\\ 
    Kemeny & \textbf{0.102} & \textbf{0.254} & \textbf{0.244} & \textbf{0.528} & \textbf{0.614}\\ 
    $k$-shortest paths & 0.280 & 1.190 & 1.965 & 5.064 & 6.343\\ 
    MPSP & 0.148 & 0.582 & 0.726 & 6.213 & 4.735\\ 
    Bidirectional & 0.104 & 0.269 & 0.272 & 0.627 & 0.631\\ 
    \hline
    \hline
    \multicolumn{6}{||c||}{Multiple UAVs with bidirectional strategy} \\
    \hline
    3 UAVs & \textbf{0.289} & \textbf{0.991} & \textbf{1.768} & \textbf{6.475} & \textbf{5.074} \\ 
    5 UAVs & 0.697 & 2.586 & 5.043 & 15.026 & 10.623 \\ 
    7 UAVs & 0.955 & 3.294 & 6.966 & 24.505 & 23.890 \\ 
     [1ex]
    \hline
    \end{tabular}
}

\label{table:avg-computation-large-20-40}
\end{table}
% \vspace{-0.6cm}

%\nn{Compared to the UGV-only strategy, focusing on the large maps, the bidirectional strategy reduces the average travel time 27\% for an equal speed ratio 20:20, 33\% for the speed ratio 20:30, and 38\% for the speed ratio 20:40. In the small maps, these values are lower 7-8\% than large maps for each speed ratio, respectively.} 
Over the 50 large maps,  the bidirectional strategy reduced the average travel time compared to the UGV-only strategy by an average of 26.7\% for an equal speed  20:20 ratio, 33.2\% for the 20:30 speed ratio, and 38.4\% for the  20:40 speed ratio. For the 50 small maps, the average reductions were approximately 7\% lower than those for the corresponding large maps at each speed ratio.
These results confirm that UAV assistance consistently improves performance, with greater benefits as the UAV becomes faster than the UGV.
% Speed 20:20: diff percentage is 7.266504615%
% Speed 20:30: diff percentage is 7.126842308%
% Speed 20:40: diff percentage is 6.981806923%
%\sa{7.0-7.3\%}
% \sa{The comprehensive results for all 50 cities in all three speed ratios will be published on GitHub.}

A key observation is that the bidirectional strategy outperformed other approaches in most maps by minimizing UGV travel times through effective UAV inspection. Incorporating multiple UAVs further reduced travel times at the expense of greater computational costs because we use the $k$-shortest paths algorithm for multiple UAVs.

\section{Addressing Practical Limitations}
\label{sec:limit}
In this paper, we focused on the \textit{replanning} component of each strategy for the DUCPP problem to evaluate how different replanning choices affect the UGV travel time. 
%To focus on this comparison, 
Towards this, we made several simplifying assumptions; relaxing them is important for practical systems.
\begin{itemize}
    \item \noindent\textit{Delay-free communication.}
% We assume UAV inspection results can be shared quickly with the UGV. 
In a disaster, network connectivity may be intermittent. However, this can be mitigated in practice by leveraging resilient communication infrastructure (e.g., satellite internet) to maintain real-time coordination.
    \item \noindent\textit{UAV energy limits.}
% Our paper currently assumes unlimited UAV energy.
%In real deployments, 
UAVs have limited battery life, and energy constraints are often a key reason to use multiple UAVs. A natural next step is to perform \textit{energy-aware} planning by assigning each UAV an energy budget and optimizing inspection routes to balance (i) expected reduction in UGV detours and (ii) energy cost, while guaranteeing that each UAV can return to a launch/recharge site. We also plan to study simple recharging strategies (e.g., selection of recharge station locations,  online UAV recharge scheduling).

    \item \noindent\textit{UAV straight-line deadheading.}
% We approximate UAV travel as straight-line motion between vertices.
UAV flights may encounter no-fly zones, obstacles,  and altitude limits. These constraints can be addressed by generating physically feasible UAV flight paths to replace straight-line travel.
%and costs 
%or by restricting UAV motion to allowed areas.
\end{itemize}

\section{Conclusion}
\label{sec:conclusion}
This paper introduced and analyzed the Dynamic UGV-UAV Cooperative Path Planning (DUCPP) problem, addressing the challenges of \saq{path finding} for a UGV in uncertain road environments. 
%
%By using UAVs to perform edge inspection,  the framework facilitates safe and efficient path finding for a UGV.
%
%%By integrating UAVs and a UGV, the framework facilitates efficient path finding through edge inspection, ensuring safe and timely \saq{path finding} for ground vehicles.
%
%were leveraged for real-time inspections,
%
UAVs can perform edge inspections to efficiently identify and eliminate impassable edges and thus reduce UGV travel times and optimize overall performance. The presented strategies, particularly the bidirectional approach, demonstrated significant improvements in reducing UGV travel times and adapting to uncertain edge conditions.  The use of multiple UAVs further reduced UGV travel times, at the cost of increased computation.

%across diverse environments and conditions
To maintain generality and ensure flexibility, we opted to use Dijkstra's algorithm for path finding. However, exploring heuristic methods such as A* or D* Lite in future work \saq{could provide valuable insights} and computational speedups.

The experimental results across diverse road networks and scenarios underscore the robustness and scalability of the DUCPP framework in environments with uncertain conditions. The findings affirm the potential of UGV-UAV cooperative path planning and inspection as a practical solution for applications such as disaster response, supply transport, and rescue missions. Future research directions include integrating heuristic path planning approaches, \saq{optimizing the assignment of multiple UAVs to subgraphs},  extending the framework to support multiple UGVs, and incorporating terrain data to estimate edge existence probabilities in challenging environments.
%\saq{considering unreliable communication networks},

\bibliography{refs.bib}

\end{document}

%\layout

\title{Dynamic UGV-UAV Cooperative Path Planning in Uncertain Environments (Supplementary Material)}
\author{Author Names Omitted for Anonymous Review. Paper-ID [680]}
\date{}

\maketitle

\section{Sample Maps}
To evaluate the effectiveness of our proposed strategies, we conducted computational experiments on a diverse set of urban road networks. Among them, we highlight five representative maps 
that illustrate different road network structures.
%\sa{that capture different structural complexities and environmental conditions.} 
These sample maps were chosen from the most populous cities across five different continents (Africa, Asia, Europe, North America, and South America) to ensure a diverse and globally representative evaluation.

For each selected city, we include two maps: a \textbf{small map} spanning approximately 1 km $\times$ 1 km and a \textbf{large map} covering 3 km $\times$ 3 km. This allows us to assess how the strategies scale across different levels of road network complexity. Figures~1 to 5 in the supplementary material illustrate both the small and large versions of each city's road network, highlighting the diversity of tested environments.

\begin{figure}[!ht]
\centering
\subfloat[Small map]{\includegraphics[width=0.45\linewidth]{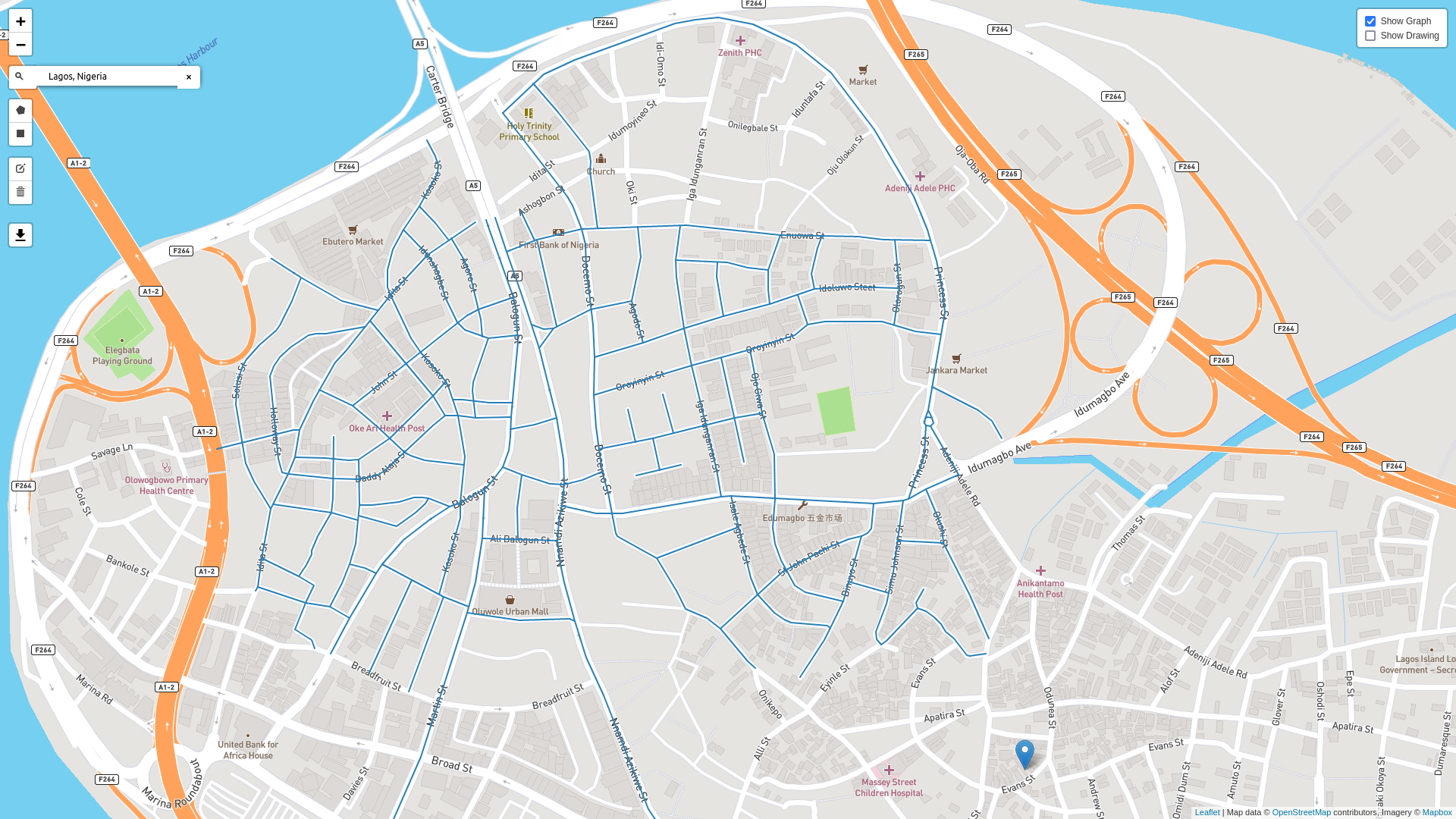}} \label{fig:lagos_a}
\hspace{10pt}
\subfloat[Large map]{\includegraphics[width=0.45\linewidth]{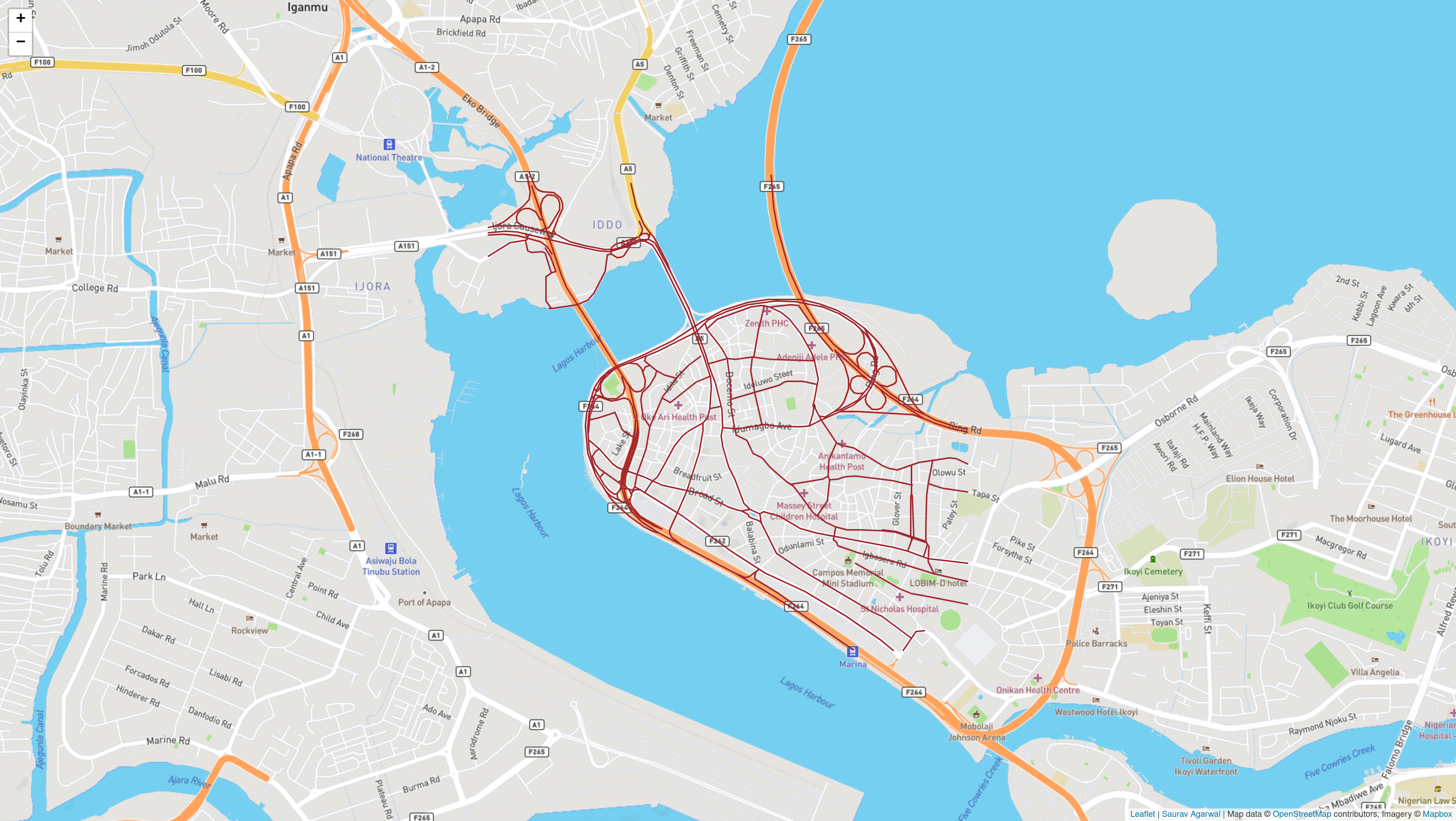}}
\label{fig:lagos_b}
\caption{Small and large maps of Lagos, Nigeria.}
\label{fig:lagos-map}
% \vspace{0.25in}
\end{figure}

\begin{figure}[!ht]
\centering
\subfloat[Small map]{\includegraphics[width=0.45\linewidth]{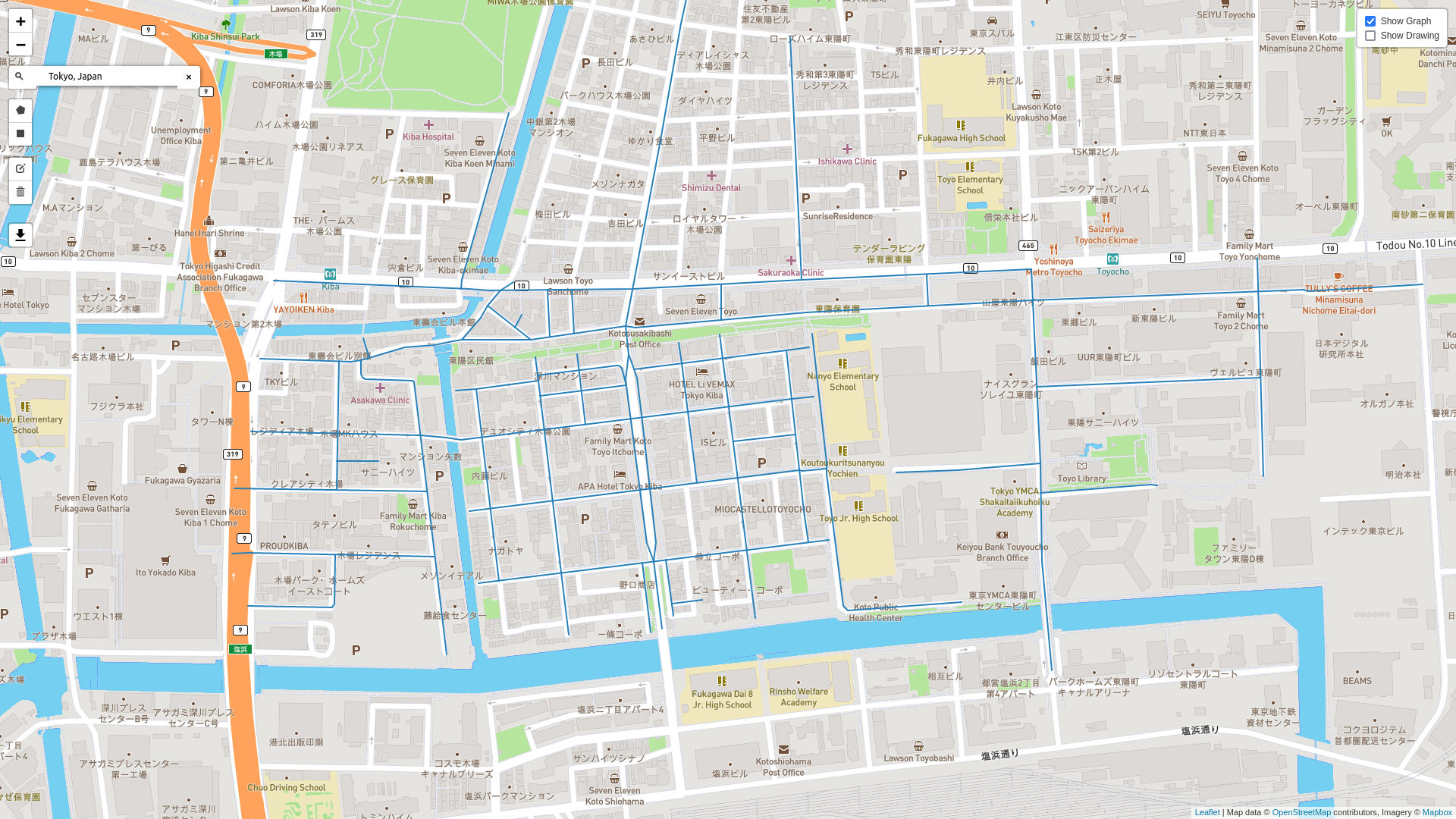}} \label{fig:tokyo_a}
\hspace{10pt}
\subfloat[Large map]{\includegraphics[width=0.45\linewidth]{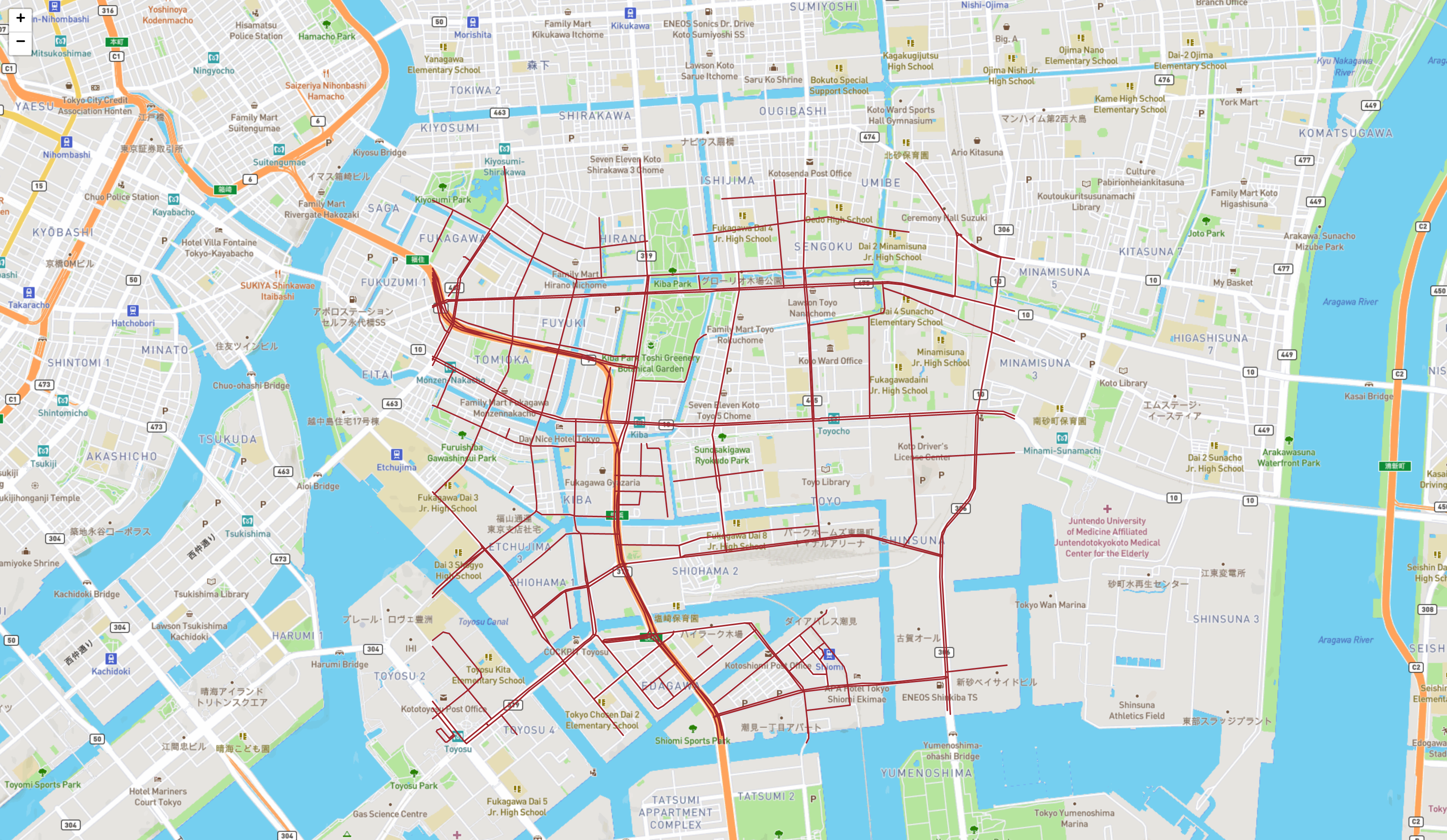}}
\label{fig:tokyo_b}
\caption{Small and large maps of Tokyo, Japan.}
\label{fig:tokyo-map}
% \vspace{0.25in}
\end{figure}

\begin{figure}[!ht]
\centering
\subfloat[Small map]{\includegraphics[width=0.45\linewidth]{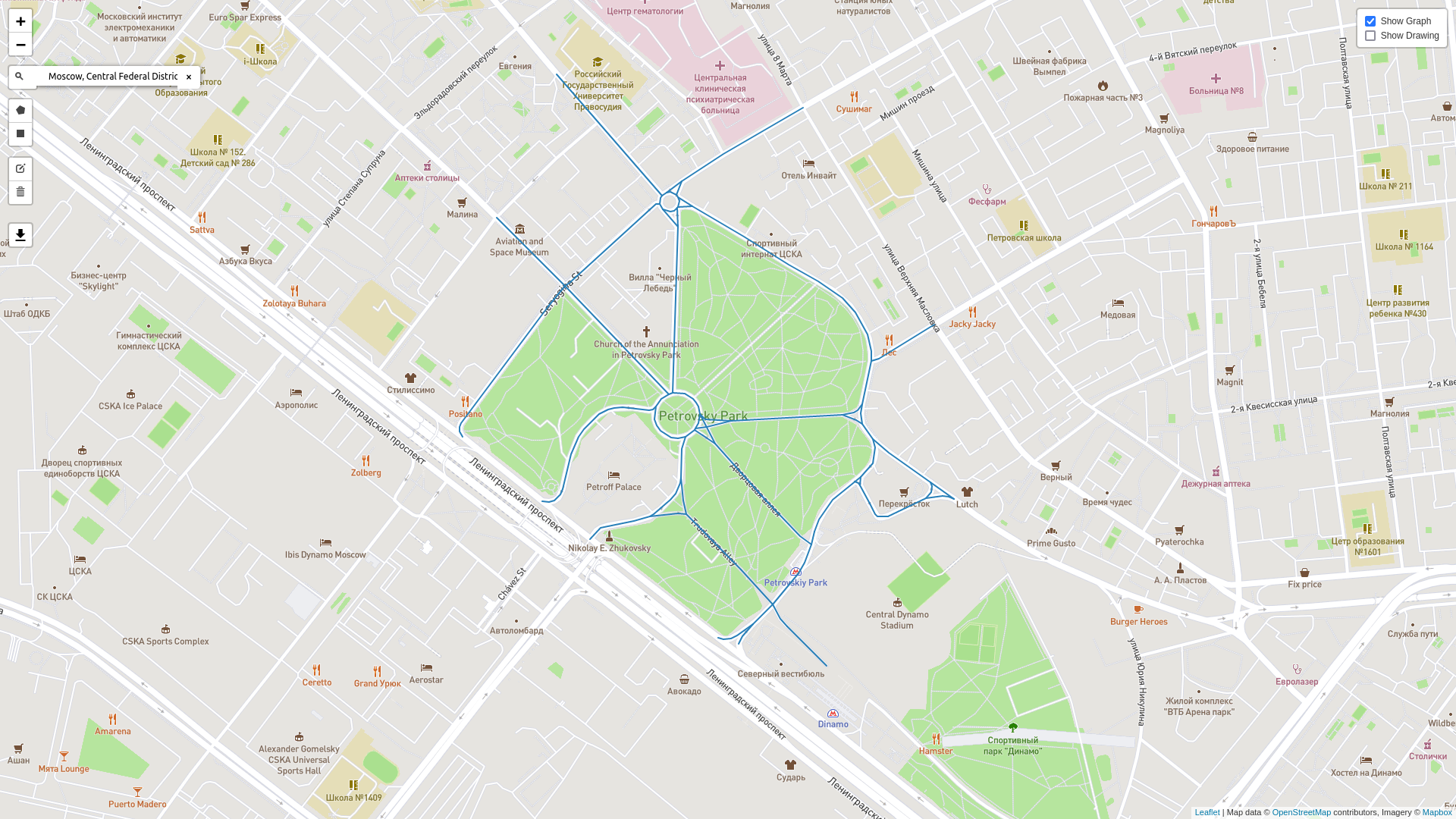}} \label{fig:moscow_a}
\hspace{10pt}
\subfloat[Large map]{\includegraphics[width=0.45\linewidth]{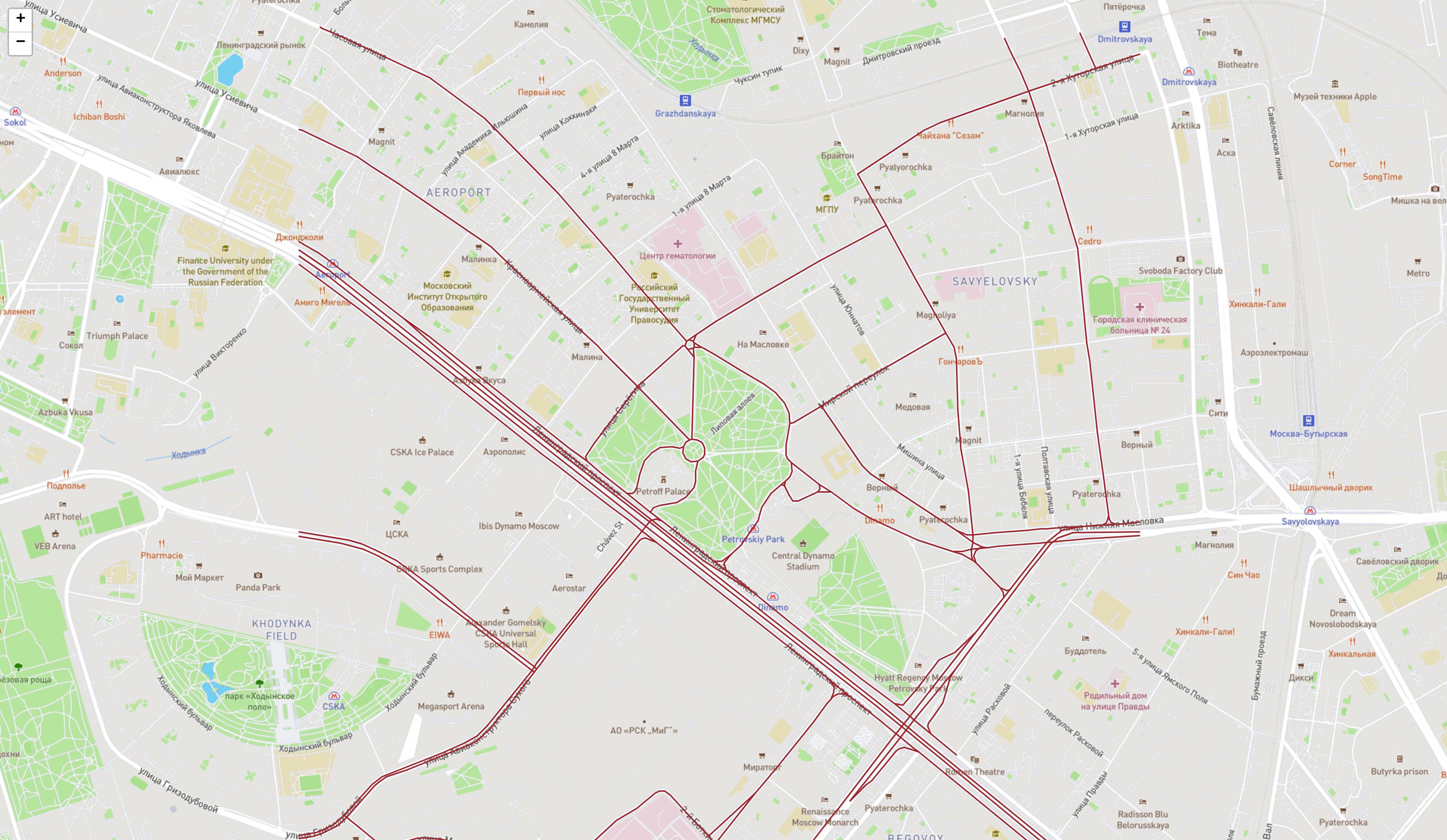}}
\label{fig:moscow_b}
\caption{Small and large maps of Moscow, Russia.}
\label{fig:moscow-map}
% \vspace{0.25in}
\end{figure}

\begin{figure}[!ht]
\centering
\subfloat[Small map]{\includegraphics[width=0.45\linewidth]{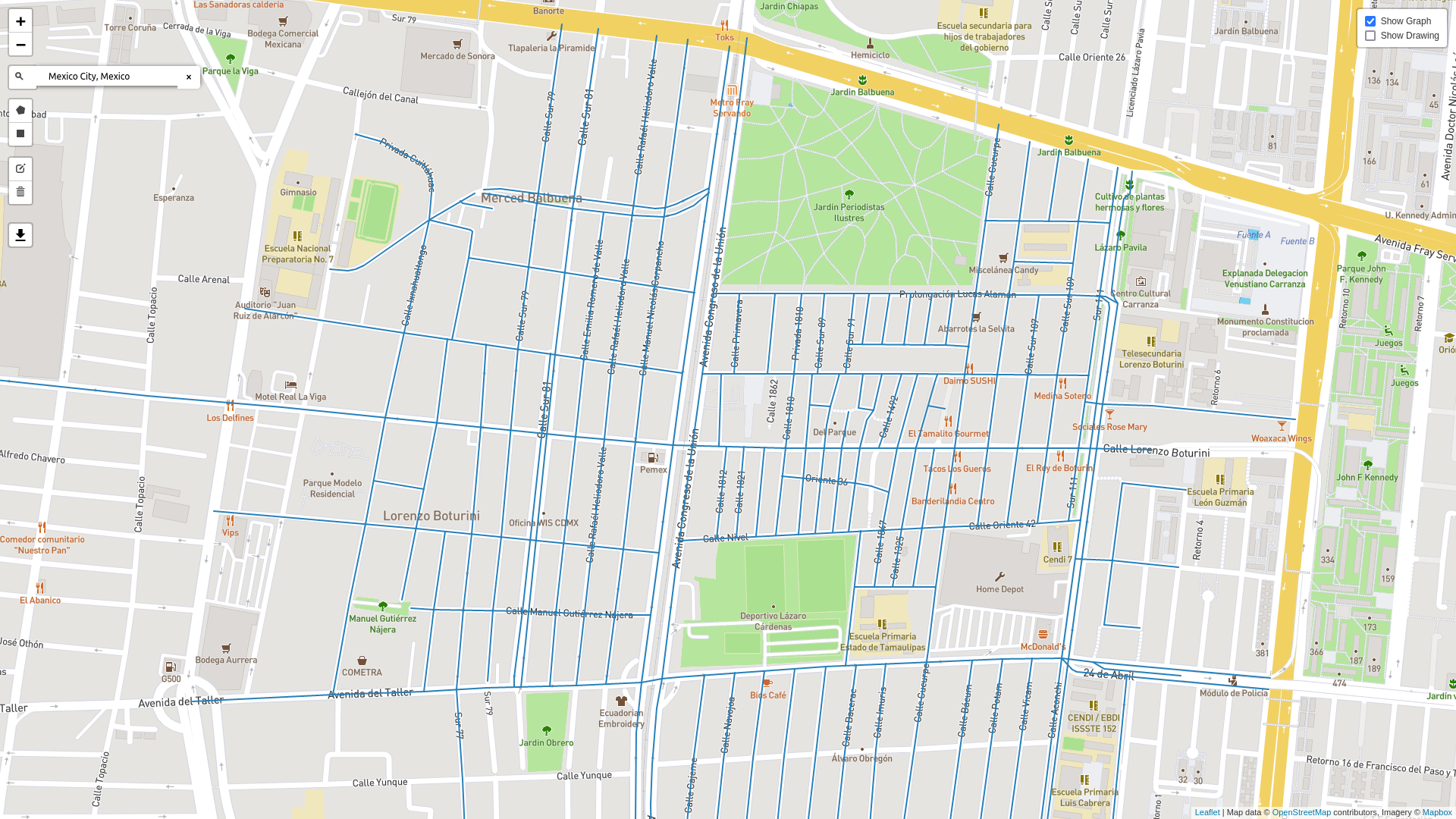}} \label{fig:mexico_a}
\hspace{10pt}
\subfloat[Large map]{\includegraphics[width=0.45\linewidth]{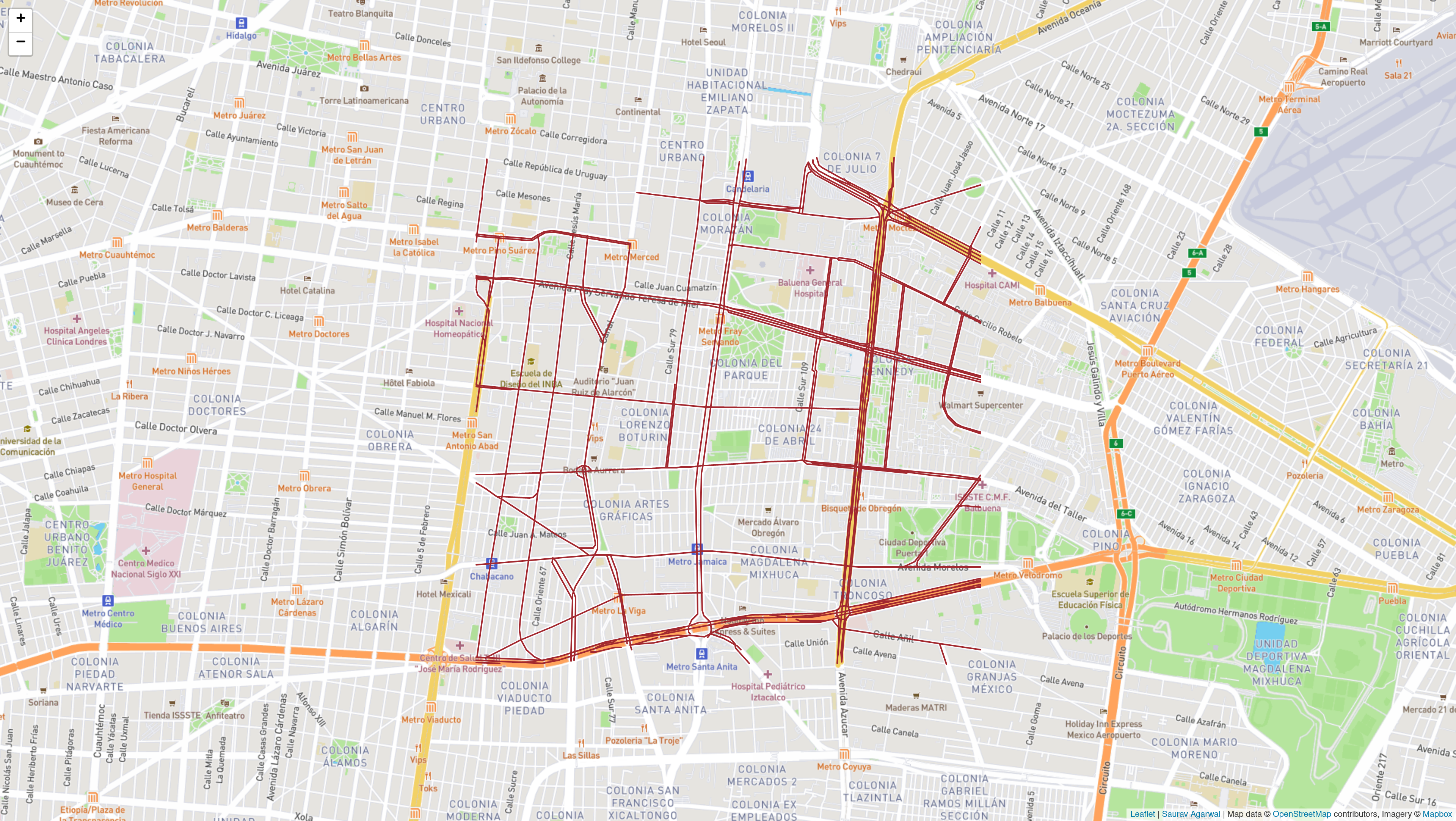}}
\label{fig:mexico_b}
\caption{Small and large maps of Mexico City, Mexico.}
\label{fig:mexico-map}
% \vspace{0.25in}
\end{figure}

\begin{figure}[!ht]
\centering
\subfloat[Small map]{\includegraphics[width=0.45\linewidth]{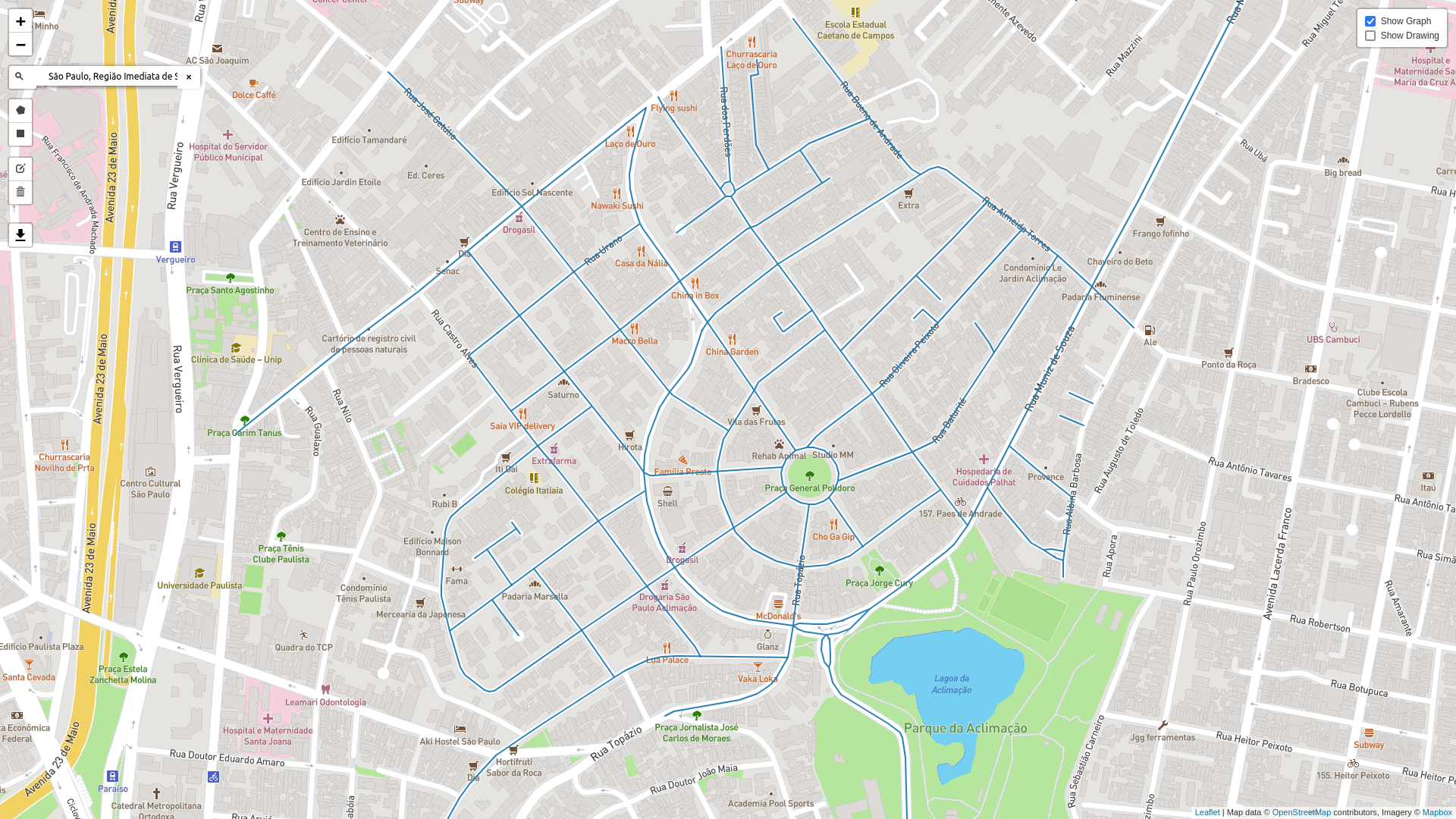}} \label{fig:sao_paulo_a}
\hspace{10pt}
\subfloat[Large map]{\includegraphics[width=0.45\linewidth]{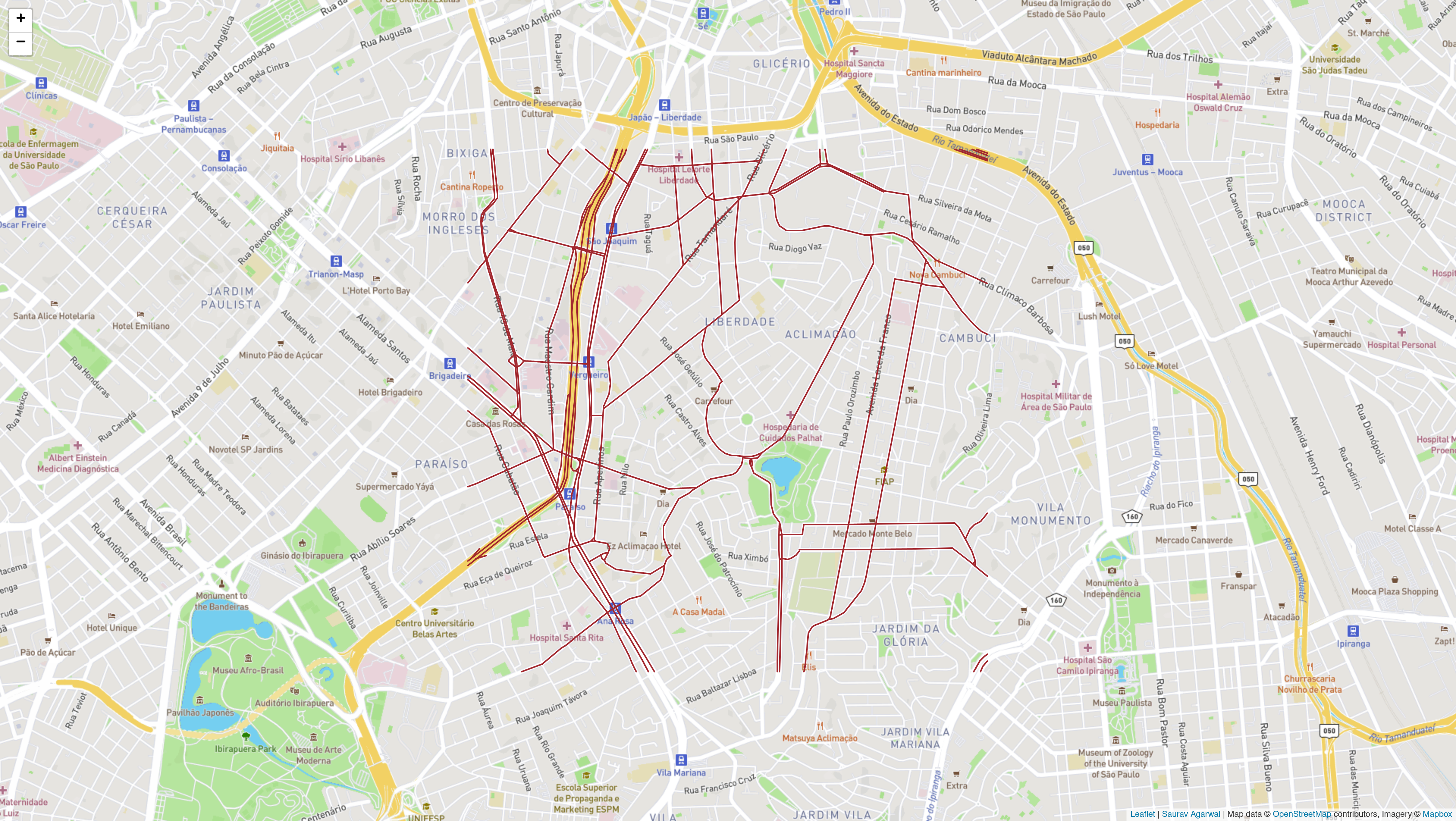}}
\label{fig:sao_paulo_b}
\caption{Small and large maps of Sao Paulo, Brazil.}
\label{fig:sao-paulo-map}
% \vspace{0.25in}
\end{figure}

%%%%%%%%%%%%%%%%%%%%%%%%%%%% TRAVEL TIME %%%%%%%%%%%%%%%%%%%%%%
\clearpage
\section{Travel Time}
This section compares the travel times of a UGV under different path-planning strategies with UAV assistance. The results highlight how UAV speed and multi-UAV coordination impact travel efficiency across various urban road networks.

%%%%% TRAVEL_SMALL_20_40
\begin{table}[!ht]
\small
\centering
\begin{tabular}{||c c c c c c | c c c||} 
 \hline
  & UGV-only & Kemeny & $k$-shortest paths & MPSP & Bidirectional & 3-UAVs & 5-UAVs & 7-UAVs \\
 \hline\hline
New York & 45.707 & 32.747 & 37.704 & 39.236 & 28.407 & 27.209 & 27.056 & \textbf{26.732} \\ 
Lima & 78.295 & 55.914 & 64.279 & 57.996 & 44.049 & 39.380 & 39.633 & \textbf{38.248} \\ 
Lahore & 82.898 & 58.329 & 65.234 & 66.987 & 54.949 & 51.065 & 50.818 & \textbf{49.583} \\ 
Paris & 74.458 & 48.405 & 55.526 & 50.609 & 39.922 & 36.171 & 36.213 & \textbf{33.693} \\ 
Mumbai & 78.461 & 51.449 & 59.220 & 50.235 & 46.765 & 43.394 & 41.775 & \textbf{39.807} \\ 
Karachi & 85.025 & 57.462 & 57.955 & 57.612 & 49.950 & 44.625 & \textbf{43.535} & 43.793 \\ 
Beijing & 107.032 & 68.224 & 84.281 & 63.696 & 56.443 & 50.463 & \textbf{49.482} & 51.052 \\
\hline
\end{tabular}
\caption{Average travel time (in seconds) for 50 small maps (UGV-UAV speed ratio 20:40).\\
The strategies are compared for a no-UAV and single UAV cases in the left part of the table. The multiple-UAV cases in the right part of the table use the bidirectional strategy.}
% \sa{Boldface the best values for single UAV case.}}
\label{table:avg-travel-small-20-40}
\end{table}

%%%%% TRAVEL_SMALL_20_30
\begin{table}[!ht]
\small
\centering
\begin{tabular}{||c c c c c c | c c c||} 
 \hline
  & UGV-only & Kemeny & $k$-shortest paths & MPSP & Bidirectional & 3-UAVs & 5-UAVs & 7-UAVs \\ [0.5ex]
 \hline\hline
New York & 45.707 & 33.712 & 41.789 & 41.195 & 29.960 & 28.688 & 28.324 & \textbf{28.079} \\ 
Lima & 78.295 & 60.750 & 71.468 & 59.359 & 46.873 & 43.723 & 43.417 & \textbf{41.393} \\ 
Lahore & 82.898 & 63.460 & 70.939 & 69.761 & 58.974 & 55.511 & 55.680 & \textbf{54.690} \\ 
Paris & 74.458 & 52.997 & 61.809 & 55.783 & 45.329 & 40.968 & 40.635 & \textbf{39.126} \\ 
Mumbai & 78.461 & 55.258 & 66.511 & 59.143 & 50.791 & 46.394 & 45.116 & \textbf{44.858} \\ 
Karachi & 85.025 & 64.032 & 65.469 & 60.435 & 55.782 & 48.344 & 48.321 & \textbf{48.154} \\ 
Beijing & 107.032 & 74.603 & 94.780 & 72.389 & 61.950 & 56.530 & \textbf{54.574} & 56.818 \\
\hline
\end{tabular}
\caption{Average travel time  (in seconds) for 50 small maps (UGV-UAV speed ratio 20:30).\\
The strategies are compared for a no-UAV and single UAV cases in the left part of the table. The multiple-UAV cases in the right part of the table use the bidirectional strategy.}
\label{table:avg-travel-small-20-30}
\end{table}

%%%%% TRAVEL_SMALL_20_20
\begin{table}[!ht]
\small
\centering
\begin{tabular}{||c c c c c c | c c c||} 
 \hline
  & UGV-only & Kemeny & $k$-shortest paths & MPSP & Bidirectional & 3-UAVs & 5-UAVs & 7-UAVs \\ [0.5ex]
 \hline\hline
New York & 45.707 & 38.296 & 44.226 & 44.078 & 33.957 & 31.908 & 31.506 & \textbf{31.305} \\ 
Lima & 78.295 & 68.831 & 75.678 & 61.499 & 54.765 & 49.669 & 52.581 & \textbf{49.239} \\ 
Lahore & 82.898 & 67.637 & 75.889 & 68.704 & 65.836 & 62.120 & 62.570 & \textbf{59.738} \\ 
Paris & 74.458 & 61.690 & 70.846 & 55.342 & 53.010 & 50.287 & 47.930 & \textbf{47.120} \\ 
Mumbai & 78.461 & 60.761 & 75.110 & 65.220 & 56.656 & 53.185 & 52.543 & \textbf{51.822} \\ 
Karachi & 85.025 & 72.232 & 79.485 & 62.212 & 64.429 & 57.365 & 54.745 & \textbf{53.518} \\ 
Beijing & 107.032 & 79.840 & 103.557 & 76.887 & 72.267 & 65.164 & 62.739 & \textbf{61.920} \\
\hline
\end{tabular}
\caption{Average travel time  (in seconds) for 50 small maps (UGV-UAV speed ratio 20:20).\\
The strategies are compared for a no-UAV and single UAV cases in the left part of the table. The multiple-UAV cases in the right part of the table use the bidirectional strategy.}
\label{table:avg-travel-small-20-20}
\end{table}

%%%%% TRAVEL_LARGE_20_40
\begin{table}[!ht]
\small
\centering
\begin{tabular}{||c c c c c c | c c c||} 
 \hline
  & UGV-only & Kemeny & $k$-shortest paths & MPSP & Bidirectional & 3-UAVs & 5-UAVs & 7-UAVs \\ [0.5ex]
 \hline\hline
New York & 219.115 & 148.077 & 168.993 & 166.567 & 115.414 & 111.185 & 110.050 & \textbf{109.898} \\ 
Lima & 386.294 & 241.958 & 262.435 & 313.092 & 187.904 & 178.460 & 174.602 & \textbf{166.065} \\ 
Lahore & 188.059 & 112.755 & 154.657 & 115.920 & 98.319 & 87.750 & 85.478 & \textbf{79.768} \\ 
Paris & 223.480 & 141.953 & 156.774 & 181.785 & 125.870 & 114.827 & 112.192 & \textbf{109.394} \\ 
Mumbai & 200.544 & 149.763 & 154.885 & 168.553 & 117.554 & 107.600 & 108.450 & \textbf{104.978} \\ 
Karachi & 271.266 & 178.327 & 226.157 & 194.211 & 156.485 & 140.857 & 143.737 & \textbf{139.021} \\ 
Beijing & 310.605 & 151.586 & 192.038 & 125.859 & 112.978 & 96.815 & 94.065 & \textbf{90.780} \\
\hline
\end{tabular}
\caption{Average travel time  (in seconds) for 50 large maps (UGV-UAV speed ratio 20:40).\\
The strategies are compared for a no-UAV and single UAV cases in the left part of the table. The multiple-UAV cases in the right part of the table use the bidirectional strategy.}
\label{table:avg-travel-large-20-40}
\end{table}

%%%%% TRAVEL_LARGE_20_30
\begin{table}[!ht]
\small
\centering
\begin{tabular}{||c c c c c c | c c c||} 
 \hline
  & UGV-only & Kemeny & $k$-shortest paths & MPSP & Bidirectional & 3-UAVs & 5-UAVs & 7-UAVs \\ [0.5ex]
 \hline\hline
New York & 219.115 & 177.699 & 213.292 & 193.765 & 133.985 & \textbf{127.690} & 128.172 & 128.096 \\ 
Lima & 386.294 & 292.732 & 369.534 & 301.111 & 243.053 & 225.363 & 222.466 & \textbf{215.421} \\ 
Lahore & 188.059 & 136.744 & 182.314 & 156.929 & 122.479 & 114.207 & 107.369 & \textbf{105.417} \\ 
Paris & 223.480 & 165.440 & 213.191 & 210.317 & 146.147 & 137.851 & 133.190 & \textbf{132.184} \\ 
Mumbai & 200.544 & 170.154 & 189.860 & 175.241 & 143.491 & 135.243 & 128.516 & \textbf{125.065} \\ 
Karachi & 271.266 & 205.980 & 270.198 & 236.504 & 189.070 & 184.252 & \textbf{178.499} & 179.208 \\ 
Beijing & 310.605 & 206.216 & 262.016 & 142.045 & 168.738 & 140.772 & 135.021 & \textbf{133.544} \\
\hline
\end{tabular}
\caption{Average travel time  (in seconds) for 50 large maps (UGV-UAV speed ratio 20:30).\\
The strategies are compared for a no-UAV and single UAV cases in the left part of the table. The multiple-UAV cases in the right part of the table use the bidirectional strategy.}
\label{table:avg-travel-large-20-30}
\end{table}

%%%%% TRAVEL_LARGE_20_20
\begin{table}[!ht]
\small
\centering
\begin{tabular}{||c c c c c c | c c c||} 
 \hline
  & UGV-only & Kemeny & $k$-shortest paths & MPSP & Bidirectional & 3-UAVs & 5-UAVs & 7-UAVs \\ [0.5ex]
 \hline\hline
New York & 219.115 & 177.699 & 213.292 & 193.765 & 133.985 & \textbf{127.690} & 128.172 & 128.096 \\ 
Lima & 386.294 & 292.732 & 369.534 & 301.111 & 243.053 & 225.363 & 222.466 & \textbf{215.421} \\ 
Lahore & 188.059 & 136.744 & 182.314 & 156.929 & 122.479 & 114.207 & 107.369 & \textbf{105.417} \\ 
Paris & 223.480 & 165.440 & 213.191 & 210.317 & 146.147 & 137.851 & 133.190 & \textbf{132.184} \\ 
Mumbai & 200.544 & 170.154 & 189.860 & 175.241 & 143.491 & 135.243 & 128.516 & \textbf{125.065} \\ 
Karachi & 271.266 & 205.980 & 270.198 & 236.504 & 189.070 & 184.252 & \textbf{178.499} & 179.208 \\ 
Beijing & 310.605 & 206.216 & 262.016 & 142.045 & 168.738 & 140.772 & 135.021 & \textbf{133.544} \\
\hline
\end{tabular}
\caption{Average travel time  (in seconds) for 50 large maps (UGV-UAV speed ratio 20:20).\\
The strategies are compared for a no-UAV and single UAV cases in the left part of the table. The multiple-UAV cases in the right part of the table use the bidirectional strategy.}
\label{table:avg-travel-large-20-20}
\end{table}

%%%%%%%%%%%%%%%%%%%%%%%%%%%% COMPUTATION TIME %%%%%%%%%%%%%%%%%%%%%%
\clearpage
\section{Computation Time}
We analyze the computation time required for different path-planning strategies, considering their feasibility for real-time applications. The results illustrate the trade-offs between computational efficiency and travel time improvements.

%%%%% COMP_SMALL_20_40
\begin{table}[!ht]
\small
\centering
\begin{tabular}{||c c c c c c | c c c||} 
 \hline
  & UGV-only & Kemeny & $k$-shortest paths & MPSP & Bidirectional & 3-UAVs & 5-UAVs & 7-UAVs \\ [0.5ex]
 \hline\hline
New York & 0.044 & 0.084 & 0.179 & 0.170 & \textbf{0.049} & 0.127 & 0.232 & 0.344 \\ 
Lima & 0.064 & 0.173 & 0.960 & 0.486 & \textbf{0.098} & 0.657 & 1.343 & 1.986 \\ 
Lahore & 0.165 & 0.383 & 3.895 & 2.262 & \textbf{0.324} & 3.609 & 7.786 & 12.392 \\ 
Paris & 0.076 & 0.188 & 1.221 & 0.514 & \textbf{0.116} & 0.807 & 1.713 & 2.388 \\ 
Mumbai & 0.052 & 0.144 & 0.995 & 0.386 & \textbf{0.088} & 0.887 & 1.751 & 2.699 \\ 
Karachi & 0.031 & 0.120 & 0.716 & 0.351 & \textbf{0.068} & 0.681 & 1.372 & 2.390 \\ 
Beijing & 0.020 & 0.070 & 0.262 & 0.106 & \textbf{0.027} & 0.139 & 0.280 & 0.481 \\
\hline
\end{tabular}
\caption{Average computation time  (in seconds) for 50 small maps (UGV-UAV speed ratio 20:40).\\
The strategies are compared for a no-UAV and single UAV cases in the left part of the table. The multiple-UAV cases in the right part of the table use the bidirectional strategy.}
\label{table:avg-computation-small-20-40}
\end{table}

%%%%% COMP_SMALL_20_30
\begin{table}[!ht]
\small
\centering
\begin{tabular}{||c c c c c c | c c c||} 
 \hline
  & UGV-only & Kemeny & $k$-shortest paths & MPSP & Bidirectional & 3-UAVs & 5-UAVs & 7-UAVs \\ [0.5ex]
 \hline\hline
New York & 0.047 & 0.088 & 0.147 & 0.118 & \textbf{0.051} & 0.107 & 0.174 & 0.268 \\ 
Lima & 0.067 & 0.165 & 0.719 & 0.312 & \textbf{0.090} & 0.489 & 1.185 & 1.716 \\ 
Lahore & 0.168 & 0.356 & 2.686 & 1.215 & \textbf{0.256} & 2.472 & 5.115 & 8.444 \\ 
Paris & 0.081 & 0.183 & 0.992 & 0.319 & \textbf{0.106} & 0.716 & 1.392 & 1.982 \\ 
Mumbai & 0.054 & 0.202 & 0.764 & 0.254 & \textbf{0.074} & 0.641 & 1.310 & 2.099 \\ 
Karachi & 0.032 & 0.145 & 0.635 & 0.232 & \textbf{0.061} & 0.540 & 1.054 & 1.677 \\ 
Beijing & 0.020 & 0.060 & 0.194 & 0.078 & \textbf{0.025} & 0.115 & 0.220 & 0.352 \\
\hline
\end{tabular}
\caption{Average computation time  (in seconds) for 50 small maps (UGV-UAV speed ratio 20:30).\\
The strategies are compared for a no-UAV and single UAV cases in the left part of the table. The multiple-UAV cases in the right part of the table use the bidirectional strategy.}
\label{table:avg-computation-small-20-30}
\end{table}

%%%%% COMP_SMALL_20_20
\begin{table}[!ht]
\small
\centering
\begin{tabular}{||c c c c c c | c c c||} 
 \hline
  & UGV-only & Kemeny & $k$-shortest paths & MPSP & Bidirectional & 3-UAVs & 5-UAVs & 7-UAVs \\ [0.5ex]
 \hline\hline
New York & 0.043 & 0.076 & 0.140 & 0.113 & \textbf{0.047} & 0.100 & 0.166 & 0.253 \\ 
Lima & 0.065 & 0.157 & 0.695 & 0.294 & \textbf{0.086} & 0.496 & 1.144 & 1.663 \\ 
Lahore & 0.167 & 0.349 & 2.559 & 1.220 & \textbf{0.252} & 2.466 & 5.022 & 8.215 \\ 
Paris & 0.076 & 0.183 & 0.957 & 0.301 & \textbf{0.103} & 0.728 & 1.356 & 1.977 \\ 
Mumbai & 0.051 & 0.139 & 0.733 & 0.251 & \textbf{0.073} & 0.629 & 1.286 & 2.029 \\ 
Karachi & 0.031 & 0.110 & 0.603 & 0.210 & \textbf{0.055} & 0.532 & 1.031 & 1.651 \\ 
Beijing & 0.020 & 0.059 & 0.188 & 0.077 & \textbf{0.025} & 0.115 & 0.216 & 0.346 \\
\hline
\end{tabular}
\caption{Average computation time  (in seconds) for 50 small maps (UGV-UAV speed ratio 20:20).\\
The strategies are compared for a no-UAV and single UAV cases in the left part of the table. The multiple-UAV cases in the right part of the table use the bidirectional strategy.}
\label{table:avg-computation-small-20-20}
\end{table}

%%%%% COMP_LARGE_20_40
\begin{table}[!ht]
\small
\centering
\begin{tabular}{||c c c c c c | c c c||} 
 \hline
  & UGV-only & Kemeny & $k$-shortest paths & MPSP & Bidirectional & 3-UAVs & 5-UAVs & 7-UAVs \\ [0.5ex]
 \hline\hline
New York & 1.110 & 1.234 & 2.623 & 2.601 & \textbf{1.140} & 1.873 & 2.859 & 4.079 \\ 
Lima & 2.098 & \textbf{2.960} & 43.490 & 30.443 & 3.136 & 30.422 & 63.426 & 96.396 \\ 
Lahore & 0.575 & 0.736 & 4.233 & 1.227 & \textbf{0.648} & 2.276 & 4.219 & 5.881 \\ 
Paris & 1.064 & 1.269 & 5.631 & 3.918 & \textbf{1.190} & 3.821 & 6.829 & 9.973 \\ 
Mumbai & 0.369 & \textbf{0.681} & 10.855 & 4.831 & 0.690 & 7.862 & 16.656 & 28.042 \\ 
Karachi & 0.294 & 0.463 & 3.779 & 1.940 & \textbf{0.389} & 2.405 & 4.728 & 7.643 \\ 
Beijing & 0.310 & 0.508 & 6.737 & 0.918 & \textbf{0.409} & 2.869 & 6.020 & 8.753 \\
\hline
\end{tabular}
\caption{Average computation time  (in seconds) for 50 large maps (UGV-UAV speed ratio 20:40).\\
The strategies are compared for a no-UAV and single UAV cases in the left part of the table. The multiple-UAV cases in the right part of the table use the bidirectional strategy.}
\label{table:avg-computation-large-20-40}
\end{table}

%%%%% COMP_LARGE_20_30
\begin{table}[!ht]
\small
\centering
\begin{tabular}{||c c c c c c | c c c||} 
 \hline
  & UGV-only & Kemeny & $k$-shortest paths & MPSP & Bidirectional & 3-UAVs & 5-UAVs & 7-UAVs \\ [0.5ex]
 \hline\hline
New York & 1.120 & 1.240 & 2.283 & 1.855 & \textbf{1.122} & 1.682 & 2.471 & 3.452 \\ 
Lima & 2.107 & \textbf{2.697} & 36.608 & 17.142 & 2.700 & 20.275 & 41.185 & 72.673 \\ 
Lahore & 0.582 & 0.716 & 2.671 & 1.039 & \textbf{0.619} & 1.887 & 3.490 & 5.193 \\ 
Paris & 1.076 & 1.251 & 4.849 & 2.749 & \textbf{1.136} & 2.987 & 5.303 & 7.620 \\ 
Mumbai & 0.369 & 0.608 & 6.857 & 2.851 & \textbf{0.572} & 6.440 & 12.242 & 18.309 \\ 
Karachi & 0.296 & 0.427 & 2.642 & 1.236 & \textbf{0.358} & 1.913 & 3.640 & 5.945 \\ 
Beijing & 0.312 & 0.484 & 4.691 & 0.630 & \textbf{0.384} & 2.570 & 5.036 & 7.978 \\
\hline
\end{tabular}
\caption{Average computation time  (in seconds) for 50 large maps (UGV-UAV speed ratio 20:30).\\
The strategies are compared for a no-UAV and single UAV cases in the left part of the table. The multiple-UAV cases in the right part of the table use the bidirectional strategy.}
\label{table:avg-computation-large-20-30}
\end{table}

%%%%% COMP_LARGE_20_20
\begin{table}[!ht]
\small
\centering
\begin{tabular}{||c c c c c c | c c c||} 
 \hline
  & UGV-only & Kemeny & $k$-shortest paths & MPSP & Bidirectional & 3-UAVs & 5-UAVs & 7-UAVs \\ [0.5ex]
 \hline\hline
New York & 1.115 & 1.239 & 2.281 & 1.865 & \textbf{1.112} & 1.679 & 2.496 & 3.466 \\ 
Lima & 2.089 & 2.705 & 36.591 & 17.144 & \textbf{2.694} & 20.258 & 41.072 & 73.285 \\ 
Lahore & 0.583 & 0.719 & 2.664 & 1.027 & \textbf{0.619} & 1.901 & 3.483 & 5.135 \\ 
Paris & 1.077 & 1.238 & 4.808 & 2.749 & \textbf{1.136} & 2.998 & 5.325 & 7.582 \\ 
Mumbai & 0.369 & 0.614 & 6.869 & 2.869 & \textbf{0.574} & 6.461 & 12.212 & 18.252 \\ 
Karachi & 0.293 & 0.428 & 2.645 & 1.230 & \textbf{0.358} & 1.918 & 3.628 & 5.929 \\ 
Beijing & 0.313 & 0.488 & 4.701 & 0.633 & \textbf{0.384} & 2.580 & 5.014 & 7.987 \\
\hline
\end{tabular}
\caption{Average computation time  (in seconds) for 50 large maps (UGV-UAV speed ratio 20:20).\\
The strategies are compared for a no-UAV and single UAV cases in the left part of the table. The multiple-UAV cases in the right part of the table use the bidirectional strategy.}
\label{table:avg-computation-large-20-20}
\end{table}

\clearpage
\section{Time Complexity}
Table \ref{table:complexity} summarizes the time complexity of all the strategies. The complexity depends primarily on the number of vertices ($V$) and edges ($E$) in the graph. The UGV-only and bidirectional strategies rely on Dijkstra's algorithm, ensuring efficient path computation. In contrast, strategies involving multiple UAVs and $k$-shortest paths introduce additional complexity due to the need for computing multiple shortest paths.
The MPSP strategy further scales with parameter $m$, as it performs Dijkstra’s algorithm $m$ times to sample candidate paths. Additionally, its complexity is influenced by $N$, the number of Monte Carlo (MC) runs in the Luby-Karp algorithm, contributing to the overall computational cost.

\begin{table}[!ht]
\centering
\begin{tabular}{||c c||}
\hline
Strategy & Complexity  \\ [0.5ex]
\hline & \\
UGV-Only (Dijkstra) & $O((V + E) \log V)$\\ 
Kemeny & --- \\ 
$k$-Shortest Paths ($k$ = 5) & $O(kV((V + E) \log V))$\\ 
MPSP ($m$ = 20) & $O(m(NE + V\log V + \log m))$ \\ 
Bidirectional & $O((V + E) \log V)$\\ \hline
\multicolumn{2}{||c||}{Multiple UAVs with Bidirectional Strategy} \\
3-UAVs ($k$ = 3) & $O(kV((V + E) \log V))$\\ 
5-UAVs ($k$ = 5) & $O(kV((V + E) \log V))$\\ 
7-UAVs ($k$ = 7) & $O(kV((V + E) \log V))$\\
 [1ex]
 \hline
\end{tabular}
\caption{Time complexity of different strategies. ($V$ is the number of vertices. $E$ is the number of edges, and $k$ represents the number of shortest paths computed)}
\label{table:complexity}
\end{table}